\documentclass[10pt]{article}
\usepackage[preprint]{tmlr}
\usepackage{amsmath,amssymb,booktabs,multirow}
\usepackage{graphicx}
\usepackage{float}
\usepackage{microtype}
\usepackage{subcaption}
\usepackage{cuted}
\usepackage{tikz}
\usetikzlibrary{arrows.meta,positioning,fit,calc}
\usepackage{pgfplots}
\pgfplotsset{compat=1.16}
\usepackage{enumitem}
\usepackage{url}

\newcommand{\E}{\mathbb{E}}
\title{Neural Interoperability Protocols: Specifying Compatibility Before Training}
\author{\name{Alireza Bagheri}\\
\addr{Independent Researcher}}
\usepackage{xcolor}
\definecolor{nipModelBorder}{HTML}{365B80}
\definecolor{nipModelFill}{HTML}{EAF0F8}
\definecolor{nipInterfaceBorder}{HTML}{247568}
\definecolor{nipInterfaceFill}{HTML}{E5F3F0}
\definecolor{nipAdapterBorder}{HTML}{A86429}
\definecolor{nipAdapterFill}{HTML}{FFF0DF}
\definecolor{nipTargetBorder}{HTML}{74558F}
\definecolor{nipTargetFill}{HTML}{F0EAF7}
\definecolor{nipPanelBorder}{HTML}{B8BEC5}
\definecolor{nipArrow}{HTML}{65717E}

\begin{document}
\maketitle

\begin{abstract}
Neural networks are increasingly developed as reusable components, yet their intermediate representations remain difficult to exchange across independently trained models. Existing approaches can align representations or encourage compatible features, but successful transfer between particular models does not by itself define what future implementations must satisfy to interoperate. We introduce \emph{Neural Interoperability Protocols} (NIPs), a specification-first framework that treats neural compatibility as a set of public interface obligations rather than a property inferred solely from model-pair performance. A NIP specifies an exchange boundary, the semantics and admissibility conditions of exchanged packets, and the downstream computations that recipients must support, independently of the architectures and training procedures implementing it. We investigate this distinction through controlled retrieval, vision, and continuous-memory experiments involving independently trained models, differing architectures and objectives, and later training cohorts. Matched-competence controls demonstrate that native task success does not ensure interoperability, while shared semantic supervision enables substantial direct transfer in the evaluated settings. Withheld-information experiments show that recipient models can continue computations requiring information unavailable to donors at exchange time. Partial-contract experiments and packet interventions further establish that observed compatibility is conditional on the semantics, packet populations, and recipient continuations being evaluated. Together, these results support public neural interface specifications as a framework for designing prospective component exchange and distinguishing demonstrated interoperability from stronger conformance claims.
\end{abstract}

\section{Introduction}\label{sec:intro}
Neural networks increasingly serve as building blocks of larger computational systems. A compelling next step is to make those blocks easier to develop, improve, and replace independently, while preserving their ability to work together. For example, a new encoder trained months after a downstream consumer could provide an improved upstream component without requiring the consumer to be retrained for that specific encoder. Achieving this form of modularity requires more than matching tensor shapes: the components must share an interpretation of the intermediate state and agree on what downstream computation it supports. Representation-similarity and symmetry studies characterize relationships among learned internal spaces \citep{raghu2017svcca,kornblith2019cka,godfrey2022symmetries}. Results from model stitching and cross-model activation transfer further demonstrate that representational correspondence alone does not guarantee successful functional interchangeability \citep{bansal2021revisiting,zhang2026negative}. This motivates public neural interfaces that implementations can target from the outset.

We argue that neural interoperability should be specified independently of the particular models used to demonstrate it. Two models may exchange intermediate representations successfully without establishing which properties a third, independently developed implementation must preserve to participate in the same exchange. Conversely, a shared interface specification can define those requirements before the participating models exist. Motivated by this distinction, we introduce \emph{Neural Interoperability Protocols} (NIPs), a framework for specifying and evaluating prospective neural component compatibility. A NIP describes the computational role of an exchange boundary, the shared meaning and admissibility conditions of the packets crossing it, and the obligations of producers and consumers. These requirements are published independently of any particular architecture, learning objective, or alignment recipe. A developer can train a producer to emit packets targeting the interface; another can train a consumer to interpret them; and future implementations can be evaluated against the same published requirements. Compatibility can then be tested by directly transplanting a producer's packet into a recipient's specified continuation, without pair-specific adaptation. Such evaluations establish interoperability under their tested conditions, while stronger conformance requirements concern the full range of packets and continuations permitted by the protocol.

\begin{figure}[t]
\centering
\includegraphics[width=0.88\linewidth]{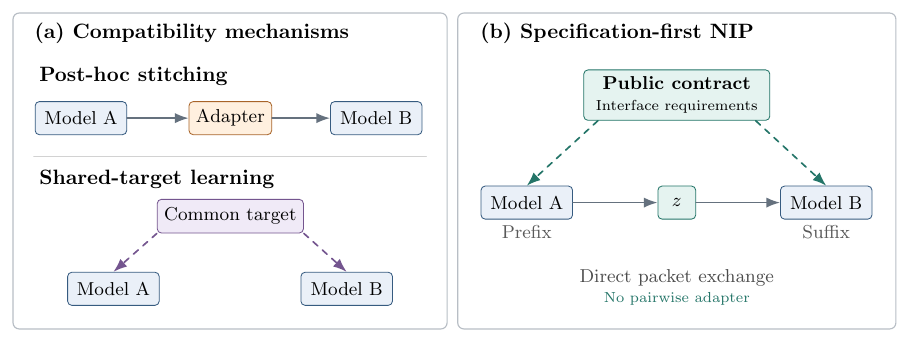}
\caption{\textbf{Compatibility mechanisms and specification-first neural interoperability.} (a) Post-hoc stitching connects pretrained components through fitted adapters, while shared-target learning encourages compatible representations through common training constraints. (b) Neural Interoperability Protocols (NIPs) define a public intermediate-state interface with explicit packet semantics, validity conditions, and producer--recipient continuation obligations. Separately trained models can target this specification and be evaluated through direct packet transplantation without pair-specific adaptation. Existing compatibility mechanisms may implement a NIP; the distinction lies in the explicit interface contract and its evaluation obligations, not in the invention of direct transfer.}
\label{fig:concept}
\end{figure}

Figure~\ref{fig:concept} contrasts the specification-first perspective of NIPs with existing compatibility mechanisms. This approach complements existing mechanisms for learning compatible features. Post-hoc stitching can connect already trained representations \citep{bansal2021revisiting}, while compatible-feature and fixed-target approaches demonstrate ways to encourage prospective reuse \citep{gygli2021reusable,biondi2024stationary,biondi2026stationary}. Such approaches can also make interface requirements explicit and may serve as mechanisms for implementing a NIP. Our emphasis is on the interface itself: specifying what must be shared, what continuation a recipient must support, and what evidence can justify a compatibility claim. The resulting framework makes prospective component exchange a concrete design and evaluation question rather than treating compatibility solely as a relationship measured between particular learned representations.

We evaluate the proposal by moving from the feasibility of direct exchange to the computational capabilities it enables, the semantic scope it standardizes, and the evidence needed to assess compatibility. We investigate these aspects through four complementary research questions (RQs). \emph{Prospective exchange (RQ1)} asks whether separately trained architectures, objectives, and later model cohorts can target a fixed interface without donor--recipient-specific adapters. \emph{Continuation-capable exchange (RQ2)} asks whether recipients can compute from transplanted intermediate states, including when essential information arrives only after the donor has produced its packet. \emph{Interface scope (RQ3)} examines which meanings and computational boundaries become reliably shared, and distinguishes task-relevant information retained privately from information made interoperable by public constraints. \emph{Compatibility evidence (RQ4)} investigates how performance on natural donor packets, alternative packet constructions, and recipient-native comparisons inform different compatibility claims. Together, these studies explore both the utility of shared neural interfaces and the importance of evaluating them against clearly stated requirements. The specific evaluation populations, prospective-freeze qualifications, and limits of packet-validity testing are detailed where their results are presented.

\noindent\textbf{Contributions.} This paper makes three connected contributions. \emph{(1) A specification-based account of neural interoperability:} we formalize compatibility in terms of public exchange boundaries, packet semantics and validity requirements, and producer--consumer obligations. This separates the interface contract from the particular architectures or training mechanisms used to satisfy it. \emph{(2) Empirical evidence for prospective and continuation-capable exchange:} we demonstrate direct intermediate-state exchange across independently trained implementations, including later-trained models and recipients that must perform downstream computations using information unavailable to donors at exchange time. \emph{(3) An evaluation framework distinguishing transfer from conformance:} through partial-contract experiments, boundary-placement studies, packet interventions, and recipient-native comparisons, we investigate what is standardized by an interface and identify the limits of the compatibility claims supported by different evaluations. Detailed relationships to earlier work appear in Section~\ref{sec:related}.

\section{Problem Setting and Neural Interoperability Protocols}
\label{sec:problem}

NIPs enable independently trained components to exchange intermediate states through a shared public interface while retaining private architectures, representations, and training procedures. This section formalizes the corresponding obligations and establishes how interoperability is evaluated.

\subsection{Public protocol and independent conformance}

\paragraph{Public contract.}
A NIP defines a public packet space $\mathcal Z$, an exchange boundary specified by computational role, packet semantics and validity requirements, and obligations for packet production and supported recipient continuations. The contract is established before participating implementations are trained, providing a shared interoperability target without prescribing their architectures, training objectives, private representations, or a unique packet for each semantic state.

\paragraph{From specification to independent implementation.}
Once the contract is published, producers can be trained to emit packets targeting its requirements, while consumers develop importers and downstream computations against the same interface without donor-specific adapters or access to eventual donors' private activations. Later implementations can target the existing protocol version and be evaluated as producers or consumers against frozen implementations. Appendix~\ref{app:nip21-profile}, Table~\ref{tab:nip21-profile}, presents NIP-4, an illustrative versioned protocol specification and compatibility-testing template, distinct from the protocols evaluated in our experiments.

\paragraph{Public exporter, private continuation.}
As one realization of the abstract protocol, a deterministic public exporter maps implementation-specific intermediate states into a common packet space, and a model-specific learned importer maps packets into states used for downstream continuation. For implementation $j$, let $a(j)$ denote its protocol-declared state-shape class, $E_{a(j)}$ the corresponding public exporter, and $I_j$ its model-specific importer. Let $h_j^{\mathrm{private}}$ denote its intermediate state, $z^{\mathrm{public}}$ the exported packet, and $\hat h_j^{\mathrm{private}}$ the imported continuation state. The within-implementation interface path is:
\begin{equation}
h^{\mathrm{private}}_j
\;\xrightarrow{\;E_{a(j)}\;}\;
z^{\mathrm{public}}
\;\xrightarrow{\;I_j\;}\;
\hat h^{\mathrm{private}}_j.
\label{eq:nip-interface}
\end{equation}
The exporter is fixed by the public specification and declared shape class rather than fitted to incumbent activations, while the importer and continuation remain model-specific. This factorization allows shared packet semantics alongside private representations; neither this decomposition nor a particular training objective is required by NIP.

\paragraph{Independence and its limits.}
A donor $i$ emits $z_i^{\mathrm{public}}=E_{a(i)}(h_i^{\mathrm{private}})$, which recipient $j$ consumes through its own importer $I_j$, without fitting a donor--recipient-specific transformation or requiring access to the donor's private activations. Whether the recipient successfully continues computation remains an empirical question: matching packet dimensions or satisfying a public semantic readout alone does not establish interoperability.

\paragraph{Training and information access.}
NIP specifies interface obligations rather than a training objective. In the compact semantic-anchoring realization, training uses
\begin{equation}
\mathcal L=\mathcal L_{\mathrm{task}}+\lambda_{\mathrm{anchor}}\mathcal L_{\mathrm{anchor}},
\label{eq:compact-anchor}
\end{equation}
where $\mathcal L_{\mathrm{task}}$ is the native-task loss, $\mathcal L_{\mathrm{anchor}}$ is the public semantic-anchoring loss, and $\lambda_{\mathrm{anchor}}\geq 0$ controls the strength of semantic supervision. Task supervision supports native computation, while semantic anchoring encourages the specified packet semantics under a fixed public readout. We examine its contribution through a no-anchor control and a changed public basis. Independently trained protocol implementations may share public data and protocol targets, but do not use one another's checkpoints, private activations, emitted packets, or pair-specific alignment losses during development. Recipients are not adapted to evaluated donors. Semantic anchoring is one possible implementation mechanism; producer validity and recipient continuation are assessed separately under the specified evaluation conditions. Appendix~\ref{app:step11-formal-detail} details the information-access rules and task-specific alternatives.

\paragraph{Validity and conformance.}
In the semantically indexed specialization formalized in Appendix~\ref{app:formal}, a task has an intermediate semantic state $s\in\mathcal S$, an allowed continuation $c\in\mathcal C$, and a required output $F(s,c)$. For each semantic state $s$, the contract defines a set $\mathcal Z_s\subseteq\mathcal Z$ of valid packets. \emph{Producer conformance} requires that every admissible input with state $s$ produces a packet in $\mathcal Z_s$. \emph{Strong consumer conformance} requires the recipient to return $F(s,c)$ for every packet $z\in\mathcal Z_s$ and every permitted continuation $c$. Together, these obligations specify sufficient conditions for pair-independent correctness.

Packet validity encompasses the complete contract-defined requirements; agreement with a public semantic decoder is sufficient only when the contract explicitly makes it so. Moreover, a formally defined validity set need not have an efficient membership test. Proposition~1 establishes the corresponding conditional consistency result under the stated conformance assumptions. When universal producer and strong-consumer conformance hold, correctness of their composition follows directly. This result establishes a formal sufficiency condition, rather than a training guarantee or empirical certification.

\paragraph{From conformance obligations to observed compatibility.}
These definitions distinguish native competence, producer packet validity, observed interoperability, and strong consumer conformance as separate aspects of compatibility assessment. Our experiments evaluate native task performance, selected packet-semantic requirements, and recipient continuation under specified model populations, packet distributions, and tasks. The vision studies additionally compare transferred packets with recipient-native performance. Appendix~\ref{app:obligation-evidence} maps these measurements to the corresponding protocol obligations and identifies the stronger conformance claims that remain unestablished.

\paragraph{Evaluating a NIP implementation.}
A practical evaluation reports the protocol version and exchange boundary, tested producer and consumer populations, packet and continuation distributions, evaluation endpoint, acceptance criterion, and observed failures. These details provide a documented basis for assessing later implementations against the same contract. Appendix~\ref{app:nip21-profile} presents an illustrative versioned reporting checklist, not an experimentally validated certification procedure. Even when packet shapes and semantic readouts match, contracts may impose different validity and continuation requirements; compatibility results must therefore be interpreted relative to the specified contract.

\subsection{Causal interoperability test}

\textbf{Testing functional exchange.} We evaluate functional exchange through \textbf{causal transplantation}: at the protocol-defined boundary, a naturally emitted donor packet replaces the recipient's intermediate packet without donor--recipient-specific transformation or recipient adaptation. The recipient then continues its computation using the transplanted state. Together with matched control interventions, this procedure tests whether the foreign packet contributes task-relevant information to downstream computation, rather than merely resembling the recipient's native representation.

\textbf{Clean and corrupted reference conditions.} To assess whether transplantation depends on task-relevant donor information, we compare the recipient's native clean computation with a matched corrupted reference and evaluate transplantation using either the donor's clean packet or its matched corrupted-control packet. These comparisons establish a recipient-relative reference and help distinguish the contribution of donor-packet information from performance attributable to other recipient inputs or the intervention procedure.

\textbf{Probability-gap normalized recovery} measures transplanted-packet performance relative to the recipient's native clean--corrupt target-probability gap. Specifically, for a clean example $x$ and matched corrupted example $\tilde{x}$, let $p_{\mathrm{clean},j,x}$ and $p_{\mathrm{corr},j,x}$ denote recipient $j$'s target probabilities on the clean and corrupted runs. For donor $i$, let $p^{\kappa}_{\mathrm{patch},i,j,x}$ denote the recipient target probability after patching either the donor's clean packet ($\kappa=\mathrm{clean}$) or its matched corrupted control packet ($\kappa=\mathrm{ctrl}$). When the recipient's clean--corrupt gap is finite and positive, we define probability-gap normalized recovery as
\begin{equation}
\rho^{\kappa}_{i,j,x}
=
\frac{
p^{\kappa}_{\mathrm{patch},i,j,x}-p_{\mathrm{corr},j,x}
}{
p_{\mathrm{clean},j,x}-p_{\mathrm{corr},j,x}
},
\qquad
\kappa\in\{\mathrm{clean},\mathrm{ctrl}\}.
\label{eq:recovery}
\end{equation}
Probability-gap normalized recovery is recipient-relative: $0$ corresponds to the corrupted recipient's target probability, and $1$ to its clean target probability. Values outside $[0,1]$ are possible when the transplanted packet performs worse than the corrupted reference or better than the clean reference. We average over evaluation examples $x$ with a finite, strictly positive recipient clean--corrupt gap; $\mathbb E_x$ denotes the empirical mean over that eligible evaluation sample. For each ordered donor--recipient pair, we obtain
\[
\rho^\kappa_{ij}
=
\mathbb E_x\!\left[\rho^\kappa_{i,j,x}\right].
\]
This probability-based metric is distinct from the normalized squared-error recovery used for continuous-task evaluation (Eq.~\eqref{eq:m20b-recovery} in Appendix D).

\textbf{Direct clean-versus-control comparison.} Probability-gap normalized recovery requires a finite, positive recipient clean--corrupt reference gap and can be sensitive to small denominators. We therefore also report \textbf{clean-versus-control advantage}, a complementary probability-space measure that compares recipient target probabilities after clean donor-packet transplantation and matched corrupted-control transplantation, without normalization by the recipient's native gap:
\begin{equation}
\Delta^{(p)}_{i,j}
=
\E_x\left[
p^{\mathrm{clean}}_{\mathrm{patch},i,j,x}
-
p^{\mathrm{ctrl}}_{\mathrm{patch},i,j,x}
\right].
\label{eq:cleancontrast}
\end{equation}
A positive advantage indicates higher recipient target probability under clean donor-packet transplantation than under the matched control, providing evidence of task-relevant donor information under the tested interventions.

\textbf{Interpreting causal interoperability evidence.} Probability-gap normalized recovery measures transplanted performance relative to the recipient's native clean--corrupt reference, whereas clean-versus-control advantage measures the target-probability difference between clean and corrupted-control donor-packet interventions. The two metrics use different reference conditions and are not numerically interchangeable. Positive effects provide evidence of task-relevant donor-packet use under the tested interventions, but do not establish strong consumer conformance over the complete protocol-valid domain.

\subsection{Prospective compatibility under a fixed protocol}

A central practical opportunity of specification-first interoperability is \emph{prospective model onboarding}: later neural components can target an existing public interface rather than develop compatibility separately with each incumbent model. To evaluate this possibility prospectively, the protocol must be fixed before entrant training, and entrants must be developed without access to incumbent-private representations or donor--recipient-specific alignment. Successful exchange is then assessed against the established protocol under specified evaluation conditions.

Let $\mathcal M_{\mathrm{inc}}$ denote the incumbent model population and $\mathcal M_{\mathrm{future}}$ a population trained subsequently from independent random initializations against the same previously fixed public protocol. The two populations may contain different model instances or architectures, but share a common protocol-defined compatibility target.

\paragraph{Clean-room information access.} Later implementations may use the published protocol, permitted public task resources, shared training datasets, and public semantic supervision. Entrant development excludes access to incumbent checkpoints, private intermediate activations, naturally emitted packets, targets derived from incumbent representations, and donor--recipient-specific alignment signals. Thus, prospective independence permits shared public resources while prohibiting incumbent-private representation access and pair-specific compatibility fitting.

\paragraph{Testing prospective interoperability.} We evaluate direct transplantation in both directions: \textbf{future$\rightarrow$incumbent}, where an incumbent recipient consumes a later model's packet, and \textbf{incumbent$\rightarrow$future}, where a later recipient consumes an incumbent model's packet. We call tested producer--recipient combinations \emph{bidirectionally compatible} when both directions satisfy their specified experimental acceptance criteria.

Successful exchange in both directions provides evidence of prospective interoperability for the evaluated models, packets, and continuations, but does not establish strong consumer conformance over the full protocol-valid domain.

\subsection{Experimental setting}
\label{sec:setup}

We begin with a controlled associative-retrieval task comprising 192 semantic cases and independently trained two-block, width-16, two-head Transformers. Additional experiments investigate variation in training objectives and architectures, prospective model onboarding, and recipient-side continuation across distinct evaluation settings. Appendix~\ref{sec:supp-experiments} details the corresponding configurations, training seeds, and intervention procedures.

\paragraph{Controlled comparison design.}
The experiments route protocol-relevant information through the public packet and the recipient's own importer, without a private donor-information bypass. We hold the public interface fixed while varying selected training mechanisms or implementation properties to examine their relationship with cross-model interoperability. Post-hoc alignment baselines are evaluated separately because they rely on different information-access assumptions.

\paragraph{Variation across implementations.}
We examine shared-interface compatibility across selected training objectives, neural architectures, and independently trained model cohorts. The vision studies separately evaluate prospective compatibility against an analytic reference, preservation of recipient-native performance, and development-stage transfer. Because these studies differ in model populations, selection histories, evaluation endpoints, and uncertainty procedures, their results are interpreted within their respective experimental designs (Appendices~\ref{app:nip21-information}, \ref{app:m20-methods}, and \ref{app:frozen-controls}).

\paragraph{Evidence and evaluation design.}
We assess native task competence and same-model packet use as qualification checks or accompanying diagnostics, reporting model-level outcomes and exceptions where applicable. We then examine cross-model exchange using complementary experimental designs: controlled ablations test the contribution of selected interoperability mechanisms; implementation-variation studies characterize transfer across evaluated training objectives and architectures; and a separate CIFAR-100~\citep{krizhevsky2009learning} cohort evaluates compatibility against a prospectively frozen analytic-reference criterion.

The CIFAR-100 evaluation specification was amended following exploratory work. An exploratory model pair preceded the final protocol freeze, and some official test identities had been encountered during earlier research. We therefore characterize this study as prospective under its documented amended freeze, rather than as untouched external validation. Uncertainty is estimated by resampling model nodes to account for repeated donor--recipient participation, with inference conditional on the selected evaluation images. The other studies retain their own model populations, endpoints, and uncertainty procedures (Appendix~\ref{sec:supp-experiments}).

These complementary evaluation settings address the four research questions, each examined through individual experimental questions (EQs). Appendix~\ref{app:obligation-evidence} maps the resulting evidence to the corresponding protocol obligations and identifies the limitations of the evaluations.

\section{RQ1: Does a Fixed Public Interface Enable Prospective Exchange?}\label{sec:rq-prospective}

\noindent\textbf{From shared semantics to prospective exchange.} RQ1 tests whether independently trained models can exchange states through a fixed public interface, examining native competence, semantic anchoring, later-model onboarding, and implementation variation. The experiments establish compatibility only for their evaluated packets and continuations.

\subsection{Native competence and semantic anchoring}\label{sec:claim-compact}
A model can perform its native task successfully without producing intermediate representations that another independently trained model can interpret. We investigate this distinction through matched-competence controls and semantic-anchoring ablations. These experiments examine how explicit public semantic requirements can coordinate intermediate representations while preserving flexibility in the underlying model implementation.

\paragraph{Matched native competence, divergent transfer.}
\noindent\textbf{EQ1:} Does equivalent native task performance imply that independently trained models can directly exchange intermediate representations, or does explicit coordination of their semantic conventions affect compatibility?

We distinguish native competence from interoperability using a two-binding associative-retrieval task: independently trained networks predict the value associated with a queried key through a four-dimensional intermediate packet. We compare four task-only models, whose packet coordinates receive no direct semantic supervision, with four semantic-head models, whose coordinates are additionally supervised against a shared four-class target. Architectures (2,816 parameters), task sources, initialization seeds, and 2,500-update budgets are matched. We test direct packet transplantation between distinct models without pairwise alignment or recipient adaptation, isolating whether equivalent native performance entails interoperable representations under these two training objectives.

All eight models achieve $100\%$ native task accuracy. Yet on 108 previously unseen structured-retrieval sources and 12 distinct directed within-arm model pairs per condition, transfer accuracy is $100\%$ for semantic-head models versus $35.65\%$ for task-only models: a prespecified $64.35$-percentage-point contrast (descriptive hierarchical bootstrap 95\% interval $[27.78,100]$ points). Cross-arm transfer is poor, as task-only recipients are not constrained to decode the public semantic convention. Thus identical native competence need not yield interchangeable packet representations; explicit semantic supervision coordinates direct exchange in this evaluated population. The additional semantic objective is not supervision-matched to the task-only condition, and this experiment establishes neither universal necessity for semantic anchoring nor superiority over prospective fixed-head alternatives (Appendix~\ref{app:matched-information}).

\paragraph{Semantic anchoring and coordinate robustness.}
\noindent\textbf{EQ2:} Is direct exchange explained by a shared low-dimensional bottleneck or coordinate system, or by supervision toward a public semantic target?

We examine whether direct interoperability arises from a shared low-dimensional bottleneck, a particular public coordinate basis, or explicit semantic supervision. Starting from a four-dimensional interface with a fixed exporter, we remove the semantic-anchor objective while preserving native task supervision, model capacity, and packet dimensionality. Although native competence is retained, the clean-minus-corrupted donor recovery advantage falls from $0.968$ (95\% confidence interval (CI) $[0.957,0.979]$) under semantic anchoring to $-0.006$ ($[-0.019,0.010]$) without it. Conversely, retaining semantic anchoring while replacing the canonical exporter with a shared dense orthonormal projection preserves strong transfer (advantage $0.993$, $[0.989,0.996]$). These controls indicate that interoperability in this realization depends on the specified semantic supervision rather than packet compactness or a privileged coordinate basis alone. They do not establish semantic anchoring as a necessary condition for interoperability in general (Appendix~\ref{app:step12-expanded-experiments}).

\subsection{Prospective vision-model interoperability}\label{sec:claim-interface}
\noindent\textbf{EQ3:} Can a later-trained architecture exchange packets with incumbent models under an unchanged public specification, without pair-specific alignment, and satisfy a prospectively frozen acceptance criterion?

We evaluate whether a fixed public interface supports interoperability across independently trained vision architectures, including models introduced after the interface specification is frozen. In an amended prospective CIFAR-100 cross-architecture evaluation, ten incumbent ResNets and ten later-trained ConvNeXts yield 200 directed cross-family packet substitutions, without pair-specific alignment, recipient adaptation, or access to incumbent-private representations during entrant training. Compatibility is assessed by the directional mean recovery gap relative to an analytic semantic reference, $G_{\mathrm{analytic}}=R_{\mathrm{analytic}}-R_{\mathrm{cross}}$, with a prespecified acceptance margin of $0.03$.

Both directions satisfy this criterion: the mean gap is $0.0212$ (95\% model-node CI $[0.0197,0.0228]$) for ResNet-to-ConvNeXt transfer and $0.0186$ ($[0.0180,0.0194]$) in the reverse direction, with both upper confidence bounds below the acceptance margin. Matched corruption controls further support the task relevance of the exchanged packets. Because the evaluation specification was amended after an exploratory pair and official-test image identities had prior research exposure, these findings support compatibility under the documented freeze rather than independent external validation. Moreover, analytic-reference compatibility does not imply preservation of recipient-native performance or conformance over all protocol-valid packets (Appendix~\ref{app:step12-expanded-experiments}, Table~\ref{tab:e4-cifar}).

\paragraph{What the frozen acceptance criterion establishes.}
The unit of the prospective decision is a \emph{directional mean} over the admitted 20-model population, not the worst donor--recipient edge and not a per-recipient replacement guarantee. Its reference is an analytic packet rather than the recipient's own emitted packet, and its $0.03$ margin was inherited from an earlier within-model tolerance rather than calibrated as a practical deployment threshold. The matched corruption contrast provides a separate causal check that clean donor state contributes information on the tested examples. Thus the result supports prospectively specified, adapter-free exchange under this particular acceptance rule; it does not assert lossless component replacement. Recipient-native effects are evaluated separately in Section~\ref{sec:recipient-native}.

\subsection{Implementation variation across objectives and architectures}\label{sec:m20-results}
A public neural interface is most useful when it can serve as a shared target across different implementation choices. We therefore examine exchange across alternative optimization objectives, independently trained later cohorts, and qualified architectures. Held-out structured retrieval tests a fixed four-dimensional public interface across two objectives, while its packet controls additionally probe what can be inferred from successful exchange on naturally produced states. Independent dSprites~\citep{dsprites17} cohorts and a qualified Transformer--GRU comparison provide complementary evidence. Design and qualification details for these implementation-variation studies appear in Appendix~\ref{app:m20-methods}.

\paragraph{Training-objective variation: structured retrieval.}
\noindent\textbf{EQ4:} Can models optimized with different training objectives directly exchange packets through the same public interface on held-out semantic sources?

Across eight independently trained implementations (four cross-entropy and four Brier), all 7,168 exchanges over 56 directed donor--recipient pairs and 128 held-out semantic sources are correct ($100\%$ exact-answer accuracy), including exchanges across the 32 cross-objective pairs (descriptive source/model-cluster bootstrap 95\% interval $[1.000,1.000]$). The mean clean-minus-wrong-target causal probability advantage is $0.9969$; wrong-target donor accuracy is $0\%$, and canonical one-hot semantic packet oracle mean accuracy is $93.75\%$. These finite-population outcomes support direct exchange under objective variation, while the saturated empirical bootstrap interval does not establish universal protocol conformance. Because the query-conditioned architecture and candidate population were selected using development-stage qualification, the evidence supports interoperability within this selected setting, not prospective generalization across arbitrary training objectives.

\paragraph{Prospective later-cohort corroboration: dSprites.}
\noindent\textbf{EQ5:} Does a previously fixed contract remain usable for bidirectional exchange when independent new model cohorts are trained later?

Under an unchanged public interface, ten incumbent models and ten independently trained later-cohort dSprites models achieve recoveries of $0.9981$, $0.9990$, and $0.9986$ for incumbent-to-future, future-to-incumbent, and future-to-future transfers, respectively. These results support bidirectional compatibility with later-trained implementations, without pairwise alignment and under the documented separation of entrant training from incumbent-private representations. The finding is specific to the evaluated cohorts and does not establish architecture-independent interoperability (Appendix~\ref{app:step12-expanded-experiments}).

\paragraph{Architecture variation: Transformer and gated recurrent unit (GRU).}
\noindent\textbf{EQ6:} Can qualified implementations from distinct sequence-model architecture families consume one another's packets under a shared interface?

We evaluate direct packet exchange between 50 Transformer and 50 two-layer GRU implementations under a shared public interface. Across 2,500 directed pairs per direction, mean clean-minus-corrupted recovery advantage is $0.832$ (95\% CI $[0.802,0.862]$) for Transformer-to-GRU transfer and $0.790$ ($[0.768,0.813]$) in the reverse direction. Although the two-layer GRU configuration met the study-level admission criterion, two of its 50 instances failed individual behavioral qualification; a separate one-layer GRU configuration also failed native-competence qualification. The results thus concern the evaluated architecture populations rather than arbitrary GRU configurations or architecture-independent conformance (Appendix~\ref{app:step12-expanded-experiments}).

\paragraph{Compatibility implications.}
Together, these studies demonstrate direct exchange without pair-specific adapters across the tested objectives, independently trained later cohorts, and qualified architecture families. They support breadth of implementation under a shared specification, while preserving the distinction between admission or qualification, observed transfer on sampled packets, and conformance over the full valid-packet domain. No result here establishes interoperability for arbitrary architectures, packet distributions, or downstream tasks.

\subsection{Recipient-native substitution under a shared interface}\label{sec:recipient-native}
\noindent\textbf{EQ7:} When a recipient processes a foreign packet rather than its own, how does recipient-relative performance change, and does that differ from passing an analytic-reference exchange criterion?

We assess recipient-native performance under direct packet substitution using a separate, image-identity-disjoint CIFAR-100 cohort comprising four ResNets and four ConvNeXts. Unlike the prospective analytic-reference evaluation, this experiment compares each recipient's recovery from foreign packets with recovery from its own intermediate states. For ResNet-to-ConvNeXt transfer, normalized squared-error recovery decreases from $0.4846$ to $0.3613$, yielding a mean recipient-native substitution gap of $+0.1233$ (95\% CI $[0.0980,0.1452]$). In the reverse direction, recovery increases from $0.3879$ to $0.4726$, yielding a gap of $-0.0847$ ($[-0.1064,-0.0645]$). All 16 cross-family pairs in each direction exhibit the corresponding sign. Matched corruption controls further support the task relevance of the exchanged packets, including when substitution degrades recipient-native recovery.

These results show that successful interface exchange need not preserve a recipient's native performance. The observed asymmetry does not distinguish donor information quality from recipient sensitivity, and the evaluation concerns a selected model population on held-out image identities. Its recipient-relative gaps are not numerically comparable with the analytic-reference gaps of the separate prospective cohort (Table~\ref{tab:m20-cifar-native}; Appendix~\ref{app:step12-expanded-experiments}).

\begin{table}[t]
\centering
\caption{\textbf{Identity-disjoint CIFAR-100 recipient-native substitution (EQ7).} Native and transfer values denote normalized squared-error recovery, not classification accuracy. The substitution gap is $g_{\mathrm{swap}}=R_{\mathrm{recipient\text{-}native}}-R_{\mathrm{transfer}}$: positive values indicate lower recovery with foreign packets. Intervals are descriptive $95\%$ three-axis bootstrap confidence intervals over donor, recipient, and shared-query axes, not guarantees for unseen model populations. This cohort and endpoint are distinct from prospective analytic-reference noninferiority.}
\label{tab:m20-cifar-native}
\small
\setlength{\tabcolsep}{4pt}
\begin{tabular}{@{}lccccc@{}}
\toprule
Direction & Pairs & \shortstack{Native\\recovery} & \shortstack{Transfer\\recovery} & $g_{\mathrm{swap}}$ & $95\%$ CI \\
\midrule
ResNet $\to$ ConvNeXt & 16 & $0.4846$ & $0.3613$ & $+0.1233$ & $[0.0980,0.1452]$ \\
ConvNeXt $\to$ ResNet & 16 & $0.3879$ & $0.4726$ & $-0.0847$ & $[-0.1064,-0.0645]$ \\
\bottomrule
\end{tabular}
\end{table}

\paragraph{Transfer beyond producer-training identities.}
To assess whether the observed transfer depends on producer-training image identities, we trained fresh ResNet-18 and ConvNeXt NIP implementations exclusively on the 45,000-image fitting partition, reserving 5,000 official-training-split images for evaluation. Frozen-recipient substitution preserved most donor-native continuation recovery in both directions: donor-relative excess MSE was $0.0119$ (95\% image-bootstrap CI $[0.0066,0.0172]$) for ResNet$\to$ConvNeXt and $0.0151$ ($[0.0127,0.0175]$) in reverse. These previously explored, fixed-checkpoint development results are exploratory, assess a different endpoint from $G_{\mathrm{analytic}}$, and cannot serve as independent confirmation of the prospective CIFAR-100 noninferiority result (Appendix~\ref{app:matched-split-nip}).

\noindent\textbf{RQ1 finding---public semantic interfaces support direct exchange across evaluated implementations.} Together, EQ1--EQ7 demonstrate direct packet exchange among independently trained implementations across the tested semantic tasks, training objectives, architecture families, and later-trained cohorts. Native task competence alone does not ensure interoperability, whereas shared semantic conventions enable exchange without pair-specific alignment in the evaluated settings. Recipient-native substitution further distinguishes successful transfer from performance-preserving component replacement. These results support specification-first interoperability for the studied models and interfaces, not universal compatibility.

\section{RQ2: Does Exchange Support Recipient-Side Computation?}\label{sec:rq-continuation}

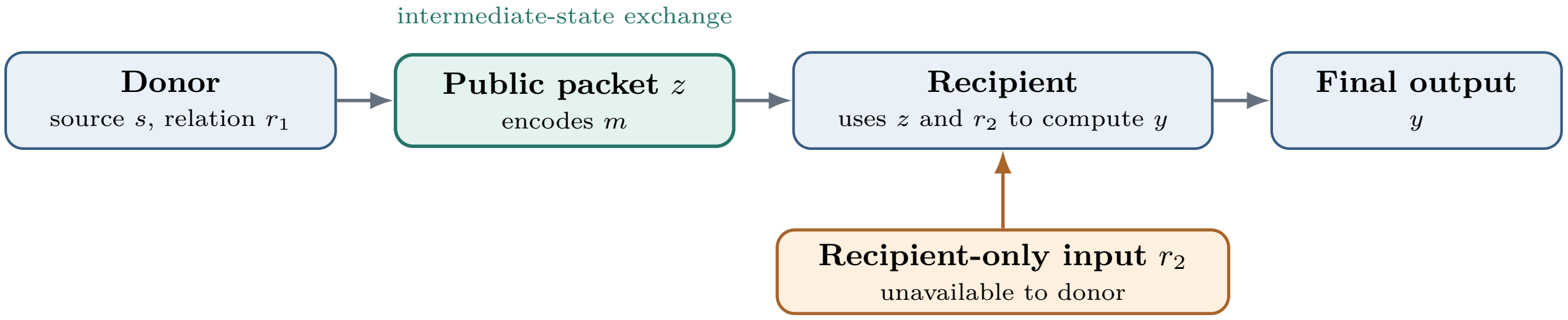
\begin{figure}[!t]
\centering
\resizebox{\linewidth}{!}{%

\begin{tikzpicture}[
  font=\small,
  >=Latex,
  node distance=5.8mm,
  model/.style={
    draw=nipModelBorder,
    fill=nipModelFill,
    rounded corners=1.8mm,
    line width=0.75pt,
    minimum height=8.5mm,
    align=center,
    inner xsep=4.5mm,
    inner ysep=1.8mm
  },
  packet/.style={
    draw=nipInterfaceBorder,
    fill=nipInterfaceFill,
    rounded corners=1.8mm,
    line width=1.0pt,
    minimum height=8.5mm,
    align=center,
    inner xsep=4.8mm,
    inner ysep=1.8mm
  },
  private/.style={
    draw=nipAdapterBorder,
    fill=nipAdapterFill,
    rounded corners=1.8mm,
    line width=0.9pt,
    minimum height=7.5mm,
    align=center,
    inner xsep=4.2mm,
    inner ysep=1.4mm
  },
  flow/.style={
    -{Latex[length=2.4mm,width=1.8mm]},
    line width=1.0pt,
    draw=nipArrow
  },
  pflow/.style={
    -{Latex[length=2.4mm,width=1.8mm]},
    line width=1.0pt,
    draw=nipAdapterBorder
  }
]

\node[model] (producer) {\textbf{Donor}\\[-1pt]\scriptsize source $s$, relation $r_1$};
\node[packet, right=of producer] (packet) {\textbf{Public packet} $z$\\[-1pt]\scriptsize encodes $m$};
\node[model, right=of packet] (recipient) {\textbf{Recipient}\\[-1pt]\scriptsize uses $z$ and $r_2$ to compute $y$};
\node[model, right=of recipient] (output) {\textbf{Final output}\\[-1pt]\scriptsize $y$};

\draw[flow] (producer) -- (packet);
\draw[flow] (packet) -- (recipient);
\draw[flow] (recipient) -- (output);

\node[private, below=8.0mm of recipient] (r2) {\textbf{Recipient-only input} $r_2$\\[-1pt]\scriptsize unavailable to donor};
\draw[pflow] (r2.north) -- (recipient.south);

\node[font=\scriptsize, text=nipInterfaceBorder, anchor=south]
  at ($(packet.north)+(0,1.3mm)$) {intermediate-state exchange};

\end{tikzpicture}}
\caption{\textbf{Testing recipient-side computation through intermediate-state exchange.} The donor receives source $s$ and relation $r_1$ and emits a packet $z$ representing the intermediate state $m=(s+r_1)\bmod 4$. The recipient receives $z$ and a separately supplied relation $r_2$, which the donor cannot observe. With uniformly sampled $r_2$, the final answer $y=(m+r_2)\bmod 4$ remains four-way ambiguous given the donor's information (optimal donor-only accuracy: $25\%$). Successful transplantation therefore requires the recipient to combine information from the transferred packet with $r_2$, rather than decode a donor-known final answer.}
\label{fig:nonterminal}
\end{figure}

A shared neural interface can support neural composition when a recipient uses another model's intermediate state to perform further computation. RQ2 examines this capability in two complementary settings. First, a withheld-information task (Figure~\ref{fig:nonterminal}) tests whether a frozen recipient can combine a foreign packet with an input unavailable when the donor produced that packet. Second, a continuous-memory task examines whether continuation-capable exchange extends to richer intermediate representations and later-trained implementations. The experiments test specified packet populations and continuations.

\subsection{Continuation with withheld information and continuous memory}\label{sec:claim-state}
\noindent\textbf{EQ8:} Can independently trained models compose a computation across an interface boundary when the donor lacks information required for the final answer?
We study a two-stage cyclic-relation task in which the donor computes an intermediate entity $m=(s+r_1)\bmod 4$ from a source $s$ and first relation $r_1$, while the recipient computes $y=(m+r_2)\bmod 4$ using a subsequently revealed relation $r_2$. Because the donor does not observe $r_2$, the final answer is uniformly distributed over four possibilities conditional on donor-visible information, imposing a donor-only accuracy ceiling of $0.25$.

Across 2,450 directed donor--recipient pairs, clean packet transplantation achieves probability-gap normalized recovery of $0.838$, compared with $0.049$ under matched corrupted-packet controls (advantage $0.789$, 95\% CI $[0.732,0.846]$). This supports recipient-side continuation from transferred intermediate states in the evaluated task, rather than transmission of a donor-known final answer. The packet may encode a sufficient intermediate statistic; the result does not establish conformance over arbitrary protocol-valid packets or continuations (Appendix~\ref{app:step12-expanded-experiments}).

\noindent\textbf{EQ9:} Does recipient-side continuation extend to sequence-valued continuous memory and later-trained implementations?
We extend the continuation setting to sequence-valued continuous memory, where a donor encodes four key--value facts into a public packet $Z\in\mathbb{R}^{4\times16}$ without observing the recipient's subsequent query. The recipient must recover the queried continuous value from the transplanted memory. Under a protocol and primary analysis fixed before training, we evaluate 30 incumbent models and 30 independently trained later implementations without pair-specific adaptation.

Across 870 directed incumbent pairs, clean-memory prediction yields mean squared error (MSE) of $0.0085$, versus $0.2995$ under a matched corrupted-memory control. The resulting control-minus-clean MSE improvement is $0.291$ (95\% CI $[0.286,0.296]$); future-to-incumbent and incumbent-to-future transfers achieve the same rounded improvement, with respective intervals $[0.287,0.296]$ and $[0.286,0.295]$. These results support recipient-side computation from naturally emitted, sequence-valued intermediate states across the evaluated cohorts, without establishing correctness for arbitrary protocol-valid memories or continuations (Appendix~\ref{app:continuous-details}).

\noindent\textbf{RQ2 finding---intermediate-state exchange supports computational composition.} Together, EQ8 and EQ9 show that independently trained recipients can use transferred intermediate representations to perform downstream computations requiring information unavailable to the donor at exchange time. This capability extends from compact semantic packets to sequence-valued continuous memories, including the evaluated later-trained implementations. The evidence establishes computational composition under the tested interfaces and continuations, rather than general consumer conformance over all protocol-valid states.

\section{RQ3: Which Meanings and Boundaries Does the Contract Standardize?}\label{sec:protocol-scope}

A public neural interface must make two aspects of component interaction explicit: \emph{where} computation is handed from one component to another, and \emph{what} information the exchanged state is intended to communicate. RQ3 examines these requirements in sequence. First, boundary-placement experiments test whether the exchange location should be defined by computational role rather than a model-specific layer address. Second, partial-contract experiments examine which semantic factors become reliably transferable when public requirements cover only part of the task state. Within-model probes and robustness controls then test whether the resulting transfer differences reflect missing information or depend on particular representation choices.

\noindent\textbf{EQ10:} Does defining an exchange boundary by computational role improve compatibility across implementation variations?\par
\paragraph{Specifying the computational boundary.}
We examine whether an exchange boundary should be specified by computational role rather than by a fixed, architecture-specific layer address. When a two-block extraction address, \texttt{B2.R1[QUERY]}, is reused in a four-block model, mean same-model recovery falls to $0.685$, reaching $0.190$ in the worst case. Selecting the corresponding final query-state boundary, \texttt{FINAL.R1[QUERY]}, instead restores reliable same-model recovery across the evaluated depth and width variants and supports cross-implementation exchange. The failure of the literal-address control, even without changing models, indicates that an interface must preserve the computational role of the exchanged state rather than its nominal architectural location. These results concern boundary specification and do not compare alternative alignment methods (Appendix~\ref{app:implementation-variation}).

\noindent\textbf{EQ11:} Does interoperability extend beyond the semantic factors explicitly standardized by a public contract?\par
\paragraph{Specifying shared semantic meaning.}
We examine whether interoperability extends to task-relevant information omitted from a public semantic contract. The experiment uses four independent, four-valued latent factors and varies the fraction explicitly anchored by shared semantic constraints. A query selecting one factor arrives only after packet emission, requiring each implementation to preserve all four factors for native task completion. At $75\%$ semantic coverage, publicly specified factors achieve $0.9532$ cross-model transfer accuracy, whereas the omitted factor transfers at $0.2563$, near the four-way chance level (Figure~\ref{fig:contract-locality}).

\paragraph{Information retention versus shared interpretation.}
To distinguish incompatible representations from information loss, we compare within-model and cross-model decoding. All four factors remain essentially perfectly recoverable within individual models, yet the omitted factor achieves only $0.252$ cross-model probe accuracy in the rotated-factor populations, compared with $0.953$ transfer accuracy for specified factors. Thus, information can remain available in independently trained representations without acquiring a shared interpretation. The evaluation covers all 256 semantic states but not the full domain of protocol-valid packets; the observed separation concerns the tested implementations and does not imply universal nontransferability of unspecified factors (Appendix~\ref{app:step12-expanded-experiments}).

\paragraph{Robustness across factor assignments and packet geometry.}
Controlled replications vary both the identities and physical positions of publicly anchored factors, while nonlinear boundary mixing tests packets beyond a simple axis-aligned arrangement. The specified--omitted transfer separation persists across these evaluated variations (Appendix~\ref{app:step12-expanded-experiments}), supporting an interpretation in terms of shared semantic constraints rather than a particular factor assignment or coordinate layout.

\noindent\textbf{RQ3 finding---interoperability is conditional on computational boundaries and specified semantics.} Together, EQ10 and EQ11 show that a public neural interface must identify both the computational role of exchanged states and the semantic obligations they are intended to satisfy. Role-defined boundaries support exchange across the evaluated implementation variations, while partial semantic contracts enable transfer of specified factors without necessarily aligning equally task-relevant information outside the contract. These findings support explicit boundary and semantic specifications while preserving freedom in implementation-private representations; they do not establish universal nontransferability of unspecified factors or conformance over the full packet-validity domain.

\section{RQ4: How Can We Evaluate the Scope of Neural Compatibility?}\label{sec:rq-evidence-boundaries}

\noindent\textbf{Evaluating the scope of observed interoperability.} A public neural interface provides a common reference for testing whether separately trained components satisfy the intended exchange requirements. RQ4 examines three complementary questions. First, does successful transfer on naturally emitted packets extend to alternative semantically interpretable packet constructions? Second, how does recipient behavior respond to representational variation that approximately preserves the public semantic readout? Third, does substituting a foreign packet preserve performance relative to the recipient's own packet? Each question concerns a different evaluation population or reference criterion. Together, they characterize observed compatibility without establishing strong consumer conformance over every protocol-valid packet and permitted continuation.

\subsection{Evaluating alternative semantic packets}
\noindent\textbf{EQ12:} Does successful transfer on naturally emitted packets extend to alternative semantically interpretable packet constructions?\par
\paragraph{Wrong-target and canonical-packet controls.}
We assess whether successful interoperability on naturally emitted packets extends to alternative semantic representations. In structured retrieval, direct transfer achieves $100\%$ accuracy on naturally emitted donor packets, compared with $93.75\%$ for canonical one-hot packets. Wrong-target donor packets yield $0\%$ accuracy, with a mean clean-minus-wrong-target causal probability advantage of $0.9969$, supporting sensitivity to the transmitted task information. The canonical-packet comparison reveals recipient dependence on packet construction despite a shared intended semantic interpretation. Because these alternative packets have not been certified against the full protocol-validity requirements, the observed discrepancy does not establish a violation of strong consumer conformance.

\subsection{Evaluating representational variation}
\noindent\textbf{EQ13:} Does agreement under a public semantic decoder imply equivalent recipient behavior across different packet constructions?\par
\paragraph{Natural and synthetic decoder-nullspace interventions.}
We examine whether agreement under a public semantic decoder implies equivalent recipient behavior. Swapping naturally emitted decoder-nullspace components between semantically matched donor packets produces negligible change in prediction error (MSE difference $-5.8\times10^{-5}$, 95\% CI $[-6.4\times10^{-4},5.1\times10^{-4}]$). By contrast, projecting packets onto the decoder row space or randomizing their nullspace components increases MSE from $0.0085$ to $0.0153$ and $0.0266$, respectively, despite changes in the decoded semantics below $9\times10^{-6}$. Thus, tolerance of naturally occurring decoder-invisible variation does not imply invariance to arbitrary decoder-equivalent packet constructions. The synthetic interventions retain positive clean-versus-corrupted recovery contrasts, indicating increased error rather than complete loss of transferred information. Because these constructed packets are not certified as protocol-valid, the observed sensitivity does not establish violations of strong consumer conformance (Appendix~\ref{app:continuous-controls}).

\subsection{Evaluating practical component substitution}
\paragraph{Recipient-native substitution in a separate identity-disjoint cohort.}
Practical component replacement requires a recipient-native comparison, not only success against an analytic reference. The identity-disjoint cohort introduced in RQ1 provides this test: swapping a foreign packet can either degrade or improve recipient-relative normalized squared-error recovery, depending on direction. Table~\ref{tab:m20-cifar-native} in RQ1 reports the native and transferred recovery, gap estimates, and descriptive intervals for both directions. As specified in Section~\ref{sec:setup}, these are neither classification accuracies nor a replication of prospective analytic-reference noninferiority.

\noindent\textbf{RQ4 finding---compatibility claims must be qualified by their evaluation scope.} Together, EQ12 and EQ13 show that successful exchange on naturally emitted packets does not establish equivalent recipient behavior under alternative semantic or decoder-equivalent representations. The recipient-native comparison in EQ7 further demonstrates that successful exchange need not preserve the recipient's own performance. These findings distinguish observed interoperability from strong consumer conformance, which requires correct behavior across all protocol-valid packets and permitted continuations. Accordingly, compatibility claims must specify the evaluated packet populations, continuation tasks, and performance references; the tested alternative packets are not independently certified as protocol-valid.

\section{Related Work}\label{sec:related}
\noindent\textbf{Positioning: a shared specification for neural compatibility.}
Research on model stitching, compatible representations, and shared latent spaces has established effective ways for neural components to exchange information and reuse learned features. NIPs build on this work by treating a public interface as an independently specified object: a boundary, packet meanings and validity sets, supported recipient continuations, and the scope of evidence needed to assess compatibility. The distinction concerns declared obligations and their evaluation, not the invention of direct transplantation, prospective reusable features, or public training targets.

\paragraph{Post-hoc alignment and reusable shared spaces.}
Model stitching recombines pretrained components through fitted representation connectors, including nonterminal recombination \citep{lenc2015equivariance,bansal2021revisiting,athanasiadis2026invariance,traft2026bridging,mai2026foundation}. Related causal-alignment methods study functionally corresponding representation subspaces \citep{grant2025alignment}. Other approaches investigate reusable shared activation interfaces \citep{kim2026activationbus} and learned latent communication between agents \citep{du2026interlat}. Our evaluated post-hoc adapters and incumbent-derived bus also achieve substantial causal transfer (Appendix~\ref{app:step12-expanded-experiments}, Table~\ref{tab:compact-mechanisms}), demonstrating that successful exchange is not exclusive to NIP training. The relevant distinction is the information used to construct an interface and whether later implementations can target it without incumbent-private access or pair-specific fitting, not method-level superiority.

\paragraph{Prospective compatible-feature learning and stationary targets.}
\citet{gygli2021reusable} establish direct recombination of compatible feature extractors and task heads, cross-architecture reuse, and prospective participation without pair-specific fine-tuning. These are precedents for direct transplantation and later component onboarding, not capabilities introduced by NIP. Stationary-representation methods develop formal compatibility and model-replacement criteria using fixed geometric targets \citep{biondi2023cores,biondi2024stationary,biondi2026stationary}, while related alignment methods use regularized transformations to preserve representation structure during model updates \citep{ricci2025lambda}. Thus public targets, shared semantic heads, and formal compatibility concepts also predate this work. NIP provides a complementary layer for specifying intermediate-state packet validity, producer requirements, recipient continuations, and the coverage of verification, while these training mechanisms may realize such a protocol. The distinction is the granularity of the stated interface and tested obligation, not evidence that prior approaches cannot satisfy equivalent requirements. We have not performed an information-matched prospective fixed-head comparison; the exploratory Gygli-inspired frozen-consumer diagnostic differs in objective, interface dimensionality, and consumer capacity and does not support a superiority claim (Appendix~\ref{app:dcc-matched}).

\paragraph{Representation similarity and functional correspondence.}
Representation-similarity and symmetry methods study relationships between learned internal spaces \citep{raghu2017svcca,kornblith2019cka,godfrey2022symmetries}. Studies of cross-model activation transfer report that representational correspondence does not consistently yield successful causal interchange \citep{zhang2026negative,piepereit2026architecture}. Related work on learned latent communication identifies private dialects and negative transfer across independently trained model societies \citep{marincat2026portable}. Studies of cross-lingual representations examine conceptual alignment and its relationship to downstream behavior, including causal evidence from activation interventions \citep{salhan2026monolingual,ravisankar2026map}. Complementary work examines representation nonuniqueness and the sensitivity of representation measurements to function-preserving reparameterizations \citep{cain2026gauge,karasan2026measurements}. In particular, prior latent-communication work reports asymmetric transfer \citep{marincat2026portable}; the identity-disjoint CIFAR-100 result here is a distinct evaluation, not the first observation of transfer asymmetry. Direct activation intervention is likewise established methodology. Our partial-contract and decoder-nullspace studies ask which properties of a public semantic readout are insufficient by themselves to ensure downstream recipient behavior; they do not imply that unstandardized factors could never be aligned.

\paragraph{Comparison by interface obligation.}
Three questions separate compatible-representation mechanisms from public-interface specification: \emph{How is the exchange space constructed?} Post-hoc connectors or buses may use incumbent representations, whereas fixed-target methods may specify a common target before implementation training. \emph{What must the recipient do?} Compatible classification or feature reuse is a valuable criterion; our withheld-information evaluation additionally tests continuation from a foreign packet and subsequently supplied information, without claiming earlier methods cannot do so. \emph{What does a passed test establish?} Native competence, semantic readout agreement, and successful transplantation on sampled natural packets differ from full producer validity and strong consumer conformance on all protocol-valid packets. These categories are not mutually exclusive: prospectively compatible features can realize a NIP if the public obligations and evaluation scope are made explicit. These distinctions position NIPs as a specification and evaluation framework that can accommodate multiple compatibility mechanisms, rather than as a competing training algorithm. The empirical scope and limitations of the resulting compatibility claims are examined in Section~\ref{sec:discussion}.

\section{Discussion and Limitations}
\label{sec:discussion}
\label{sec:limitations}

\paragraph{Prospective interoperability and neural composition.}
Our results demonstrate the feasibility of public neural interfaces as reusable development targets for independently trained components. Later-cohort exchange establishes interoperability with implementations developed after the interface was fixed, while withheld-information continuation demonstrates that exchanged representations can support downstream computation using information unavailable to the producer. Matched-competence controls further show that native task performance alone does not guarantee interoperability. Together, these findings establish the feasibility of specification-first neural composition under the evaluated contracts and information-access conditions, without implying interoperability across arbitrary models or superiority over alternative approaches.

\paragraph{From public specifications to modular development workflows.}
Beyond individual exchange demonstrations, the specification-first approach suggests a development workflow in which interface obligations remain stable while component implementations are developed and evaluated independently. Our experiments illustrate three aspects of this workflow: onboarding later implementations under a fixed specification, composing components through intermediate-state exchange, and evaluating compatibility against explicitly defined functional criteria. Together, these capabilities provide a foundation for modular neural development in which implementations can evolve independently while targeting a shared interface contract.

\paragraph{Matching evaluation criteria to intended use.}
Different compatibility evaluations measure distinct properties of a neural interface. Analytic-reference recovery assesses performance against a shared target, whereas recipient-native CIFAR-100 substitution measures the cost of replacing a recipient's own representation and reveals directional asymmetries. In structured retrieval, natural-packet exchange achieves perfect transfer on the evaluated population, yet canonical semantic packets can elicit different recipient behavior. Together, these findings demonstrate that compatibility claims must be interpreted relative to the evaluated packet population, recipient continuation task, and performance criterion, rather than reduced to a single universal score.

\paragraph{Explicit semantics and packet validity.}
Our findings highlight the importance of specifying both the semantic scope and the admissible packet population of a public neural interface. Boundary-placement experiments compare operational extraction rules without establishing a complete, executable packet-validity predicate. In the partial-factor study, publicly specified factors transfer reliably across models, whereas omitted but task-relevant factors can remain recoverable within individual models without acquiring a shared cross-model interpretation. Decoder-nullspace experiments reveal a complementary distinction: naturally emitted variation invisible to the semantic decoder is tolerated in the evaluated continuations, while selected synthetic perturbations that approximately preserve the semantic readout can change recipient behavior.

These observations delineate the scope of the evidence rather than establish failures of protocol conformance. Although the partial-factor study exhaustively evaluates all 256 semantic states, it does not cover every valid packet associated with each state. Natural-packet evaluations sample representations emitted by the tested producers, and neither canonical semantic packets nor synthetic decoder-equivalent packets have been certified as members of the corresponding valid-packet sets $\mathcal Z_s$. Consequently, differences in recipient behavior under these constructions do not establish violations of strong consumer conformance. The results instead motivate explicit semantic obligations, operational packet-validity criteria, and clearly specified recipient continuation requirements.

\paragraph{Prospective evaluation and selection boundaries.}
The prospective CIFAR-100 cross-architecture evaluation provides evidence for interoperability under a documented, amended evaluation freeze. It evaluates 200 directed substitutions using a frozen directional-mean noninferiority criterion, model-node uncertainty, and positive corruption controls. The $0.03$ acceptance margin applies specifically to the analytic-reference recovery gap, not to recipient-native substitution or the subsequent training-population-matched development analysis. Because one exploratory pair was examined before the final evaluation freeze and some official test identities had been exposed during earlier exploration, the study should be interpreted as a prospective evaluation under the amended freeze rather than as fully independent external validation.

A separate 45,000/5,000 training-population-matched analysis examines sensitivity to training--development overlap. This analysis reuses previously examined development identities, and its image-bootstrap intervals quantify uncertainty conditional on the evaluated checkpoints rather than variation across independently trained model populations. Neither evaluation establishes robustness to distribution shift or generalization across arbitrary architectures.

\paragraph{Relationship to alternative compatibility methods.}
The comparison experiments reinforce the distinction between specifying interoperability requirements and selecting methods to satisfy them. Although the evaluated training recipes differ in their behavior under frozen-consumer substitution, confounded objectives, interface dimensions, and consumer capacities preclude general claims of method-level superiority. The successful transfer achieved by alternative alignment mechanisms further supports treating NIPs as complementary to existing approaches. The central contribution is therefore not a preferred optimization procedure, but a framework for defining and evaluating compatibility obligations independently of implementation choices (Appendices~\ref{app:nip21-information} and~\ref{app:dcc-matched}).

\paragraph{Opportunities for interface evolution.}
Separating public interface specifications from their implementations creates a foundation for managing interoperability as models evolve. In principle, this separation would allow developers to distinguish implementation updates that preserve packet semantics and recipient obligations from interface changes that require explicit versioning or migration. Our experiments demonstrate later-model onboarding under fixed contracts but do not evaluate protocol evolution, version negotiation, cross-version compatibility, or the operational costs of maintaining shared standards. Establishing these capabilities, including interoperability among independently developed third-party implementations, is an important direction for future work.

\paragraph{Toward broader neural interoperability.}
Extending specification-first interoperability beyond controlled evaluations requires addressing the diversity, uncertainty, and operational demands of real-world neural systems. Important challenges include defining semantics for open-ended pretrained representations, evaluating heterogeneous implementations under distribution shift, and establishing operational conformance criteria. The evaluated model populations are finite and selected through documented admission procedures; consequently, the reported uncertainty estimates do not establish guarantees for arbitrary unseen implementations. Earlier CIFAR-100 development used the unrestricted official test split, while the held-out cross-architecture analysis sampled with replacement from its eligible image partition (Appendix~\ref{app:m20-audit}). Robustness to adversarial or information-leaking packets also remains untested. Moreover, conventional typed interfaces may be preferable when exchanged states are fully symbolic. For learned intermediate representations, promising directions include stronger recipient-side conformance tests, semantic contracts requiring less direct supervision, and operational packet-validity criteria. Addressing these challenges would extend the present evidence toward more general modular neural systems in which independently developed components interoperate under explicit public contracts.

\section{Conclusion}
\label{sec:conclusion}
Neural Interoperability Protocols (NIPs) establish a specification-first framework for defining and evaluating intermediate-state exchange independently of particular model implementations. Across controlled tasks and heterogeneous model families, our experiments demonstrate that independently trained components can exchange representations under fixed public contracts, including prospective onboarding of later-trained models and recipient-side computation using information unavailable to the donor. The results further distinguish native task competence, successful packet exchange, and strong protocol conformance as separate properties, while showing that interoperability extends only to the semantic obligations and packet populations supported by the available evidence. Together, these findings establish a foundation for modular neural systems in which compatibility is an explicit, testable interface requirement rather than an assumed consequence of shared task performance.


\section*{Broader Impact Statement}

Neural Interoperability Protocols could facilitate modular neural systems by enabling independently developed components to exchange intermediate representations through explicitly specified interfaces. Potential benefits include easier component replacement, clearer compatibility requirements, and more systematic evaluation of interoperability. However, exchanging intermediate representations may expose sensitive information, introduce security risks through untrusted packets, or encourage reliance on compatibility claims that exceed the evaluated conditions. These risks motivate appropriate access controls, operational packet-validity checks, privacy-leakage assessments, and recipient-side robustness testing before deployment. Our experiments use controlled synthetic tasks and benchmark image data and do not establish safety, security, or reliability in deployed systems. The broader impact of NIPs will depend on their applications and the safeguards adopted by implementers.

\small

\normalsize

\par\bigskip
\appendix
\section*{Guide to the Supplementary Material}
The supplementary material provides the protocol specifications, experimental evidence, methodological details, and formal foundations supporting the main paper. The appendices are organized as follows.
\begin{description}
\item[Appendix~\ref{app:obligation-evidence} --- Protocol obligations and experimental evidence.] Maps declared NIP requirements to the tests that evaluate them, distinguishing demonstrated properties from stronger conformance claims that remain unestablished.
\item[Appendix~\ref{app:step11-formal-detail} --- Interface specification and realization.] Defines the public interface and conformance obligations, explains protocol realization and information-access rules, and presents an illustrative compatibility-testing profile.
\item[Appendix~\ref{app:step12-expanded-experiments} --- Additional experimental results.] Presents supporting evidence for RQ1--RQ4, covering direct and prospective exchange, recipient-side continuation, semantic contract scope, and behavioral conformance boundaries.
\item[Appendix~\ref{sec:supp-experiments} --- Experimental protocols and supporting analyses.] Documents evaluation design, information access, experimental controls, statistical methods, reproducibility limitations, and exploratory comparisons.
\item[Appendix~\ref{app:formal} --- Formal details.] Specifies mathematical assumptions and conformance definitions, establishes conditional compatibility results, and distinguishes public-interface agreement, private representational freedom, and full protocol validity.
\end{description}
Together, these appendices distinguish the NIP specification, evidence from evaluated implementations, and stronger properties that have not been established.

\clearpage
\section{Protocol Obligations and Experimental Evidence}\label{app:obligation-evidence}
\noindent\textbf{How to read this map.} Table~\ref{tab:obligation-evidence} links each declared NIP property to a reported test, the conclusion justified by that test, and a stronger conclusion not established by the available evidence. The table is an evidence map, \emph{not} a certification report: passing finite tests on selected packets and continuations does not imply universal protocol conformance. The underlying operational requirements are specified in Appendix~\ref{app:step11-formal-detail} and formalized in Appendix~\ref{app:formal}.

\begin{center}
\begin{minipage}{\linewidth}
\captionof{table}{\textbf{Obligation--evidence map.} Each row distinguishes an interface requirement from the empirical evidence for selected implementations. ``Not established'' denotes an unsupported inference, not a demonstrated failure. The illustrative operational profile (Appendix~\ref{app:nip21-profile}) was not implemented as a certification regime.}
\label{tab:obligation-evidence}
\centering
\begingroup
\footnotesize
\setlength{\tabcolsep}{3pt}
\renewcommand{\arraystretch}{1.13}
\begin{tabular}{p{0.19\linewidth}p{0.25\linewidth}p{0.275\linewidth}p{0.22\linewidth}}
\toprule
\textbf{NIP property} & \textbf{Test / evidence} & \textbf{Demonstrated in scope} & \textbf{Not established} \\
\midrule
Public interface; separate training & Fixed packet/readout; separately initialized, access-restricted cohorts (RQ1) & Evaluated models target a shared interface without pair-specific fitting & Independence beyond the documented access restrictions \\
\addlinespace[2pt]
Producer semantics & Held-out public-readout and task-specific admission checks (RQ1) & Sampled agreement with checked semantics & Producer validity over all admissible source inputs \\
\addlinespace[2pt]
Recipient continuation & Frozen-recipient transplantation on specified inputs and continuations (RQ1--RQ2) & Foreign natural packets support the tested continuations & Correctness for every $z\in\mathcal Z_s$ and permitted $c$ \\
\addlinespace[2pt]
Semantic contract locality & Partial-anchor study across all 256 factor states (RQ3) & Anchored factors transfer; omitted factors need not & Exhaustive validity-domain coverage \\
\addlinespace[2pt]
Alternative packet constructions & Canonical retrieval packets; decoder row-space and nullspace interventions (RQ4) & Natural-packet success or decoder consistency does not guarantee success on altered packets & Altered-packet validity or failure of strong conformance \\
\addlinespace[2pt]
Natural decoder-null variation & Matched, naturally emitted null-component swaps (RQ4) & Tolerance under tested continuations & Invariance for all decoder-equivalent packets \\
\addlinespace[2pt]
Packet causal contribution & Matched clean/corrupted controls (RQ1--RQ2; interpreted in RQ4) & Task-relevant benefit under reported endpoints & Universal semantic or behavioral correctness \\
\addlinespace[2pt]
Prospective compatibility & Qualified later/incumbent cohorts under fixed contracts (RQ1) & Reported prospective exchange criteria met in evaluated cohorts & Arbitrary future-model or distribution-wide exchange \\
\addlinespace[2pt]
Strong consumer conformance & Formal universal obligation; no exhaustive empirical test & A defined ideal requirement for independent recipients & Empirical certification over all valid packets and continuations \\
\addlinespace[2pt]
Invalid packets and versioning & Illustrative profile only (Appendix~\ref{app:nip21-profile}) & Proposed operational checks & Implemented rejection or negotiation guarantees \\
\bottomrule
\end{tabular}
\endgroup
\end{minipage}
\end{center}

\noindent\textbf{Interpretation.} The reported evidence supports direct exchange, recipient-side computation, and coordination of specified semantics within the evaluated populations. Broader claims require additional validity-domain, continuation, independent-implementation, and operational-version tests. The formal obligations and experimental qualifications are detailed in Appendices~\ref{app:formal} and~\ref{sec:supp-experiments}, respectively.

\clearpage
\section{Extended Interface Specification and Realization}\label{app:step11-formal-detail}
\subsection{Interface Specification and Conformance Obligations}\label{app:nip-specification}
\paragraph{Neural Interoperability Protocol.}
A NIP defines a public interface contract that separately trained neural components can implement without access to one another's private representations. The contract is published before implementations are developed or adapted to target it and defines:
\begin{enumerate}[leftmargin=1.4em]
    \item a public packet space $\mathcal Z$ for exchanging intermediate information;
    \item a computational boundary specifying the role of the exchanged state;
    \item public semantic and validity requirements for packets at that boundary; and
    \item obligations for packet production, exchange, and supported recipient continuations.
\end{enumerate}
The contract provides a common interoperability target without prescribing private architectures, representations, optimization methods, or a particular semantic-anchoring mechanism. NIP thus distinguishes the public interface specification and its compatibility obligations from the implementation mechanisms used to satisfy them.

\paragraph{From specification to implementation and evidence.}
NIP distinguishes three complementary activities. \emph{Specification} defines the public packet space, computational boundary, semantic and validity requirements, permitted continuations, and exchange rules independently of participating models' private representations. \emph{Realization} develops components against this published contract, selecting architectures, exporters, importers, training objectives, and supervision strategies. Fixed semantic readouts and semantic-anchoring losses are possible realization mechanisms, not requirements of NIP. \emph{Evaluation} examines producer-side packet properties, direct exchange, and recipient continuation under specified conditions, including controlled interventions where appropriate.

This separation allows implementations with different architectures and training methods to target the same public interface while keeping their evaluation obligations explicit. Finite tests provide evidence for the evaluated implementations, packets, and continuations; they do not establish conformance over the complete protocol-valid domain.

\noindent\textbf{Distinct obligations and tests.} Compatibility assessment distinguishes native task competence, producer-side packet checks, direct donor-packet transplantation, and strong consumer conformance. Native evaluation tests a component's own task performance; producer-side checks examine selected packet requirements; and donor transplantation tests recipient continuation using packets naturally emitted by independently trained implementations. Strong consumer conformance instead requires correct continuation for every protocol-valid packet and permitted continuation, as formalized in Appendix~\ref{app:formal}. Finite experimental success does not establish this universal obligation.

A public semantic decoder may specify or test selected packet requirements, but decoder agreement establishes complete packet validity only when the contract explicitly makes that check sufficient. Additional validity constraints may apply, and packets sharing a semantic readout need not yield equivalent recipient behavior. The protocol-valid set is therefore defined by the contract, not by the packets observed during evaluation.

\paragraph{Operational admission and conformance.}
Operational admission assesses whether a candidate implementation satisfies specified interface checks under declared evaluation conditions. Admission decisions depend on the tested model and packet populations, diagnostics, and acceptance criteria; they do not establish universal producer or strong consumer conformance. In particular, passing producer-side semantic checks and native-recipient qualification tests does not guarantee successful continuation after foreign-packet substitution.

The structured cross-objective cohort demonstrates successful direct transplantation of naturally emitted packets under its evaluated conditions. In contrast, the separate identity-disjoint image cohort exhibits direction-dependent substitution gaps despite model admission. These findings motivate separate reporting of producer-side packet checks and recipient-side continuation outcomes. Neither study establishes correctness over the entire contract-defined validity domain $\mathcal Z_s$.

\paragraph{Worked example: a semantically indexed NIP-4 interface.}
Consider a task with an intermediate semantic state $s\in\mathcal S$, a permitted continuation input $c$, and required downstream output $F(s,c)$. An illustrative NIP-4 contract specifies the computational boundary, public packet space $\mathcal Z=\mathbb R^4$, semantic and validity requirements for each state, and identity transplantation $\tau(z)=z$, under which the public packet is exchanged unchanged.

Participating implementations independently develop their upstream computations and model-specific importers against the published contract, without donor--recipient-specific fitting. Operational evaluation examines whether producers satisfy specified packet checks on held-out inputs and whether recipients continue successfully when their native packets are replaced by naturally emitted packets from independently trained donors, with continuation $c$ supplied.

Passing these tests provides evidence for the evaluated packets and continuations, not correctness over the complete protocol-valid sets $\mathcal Z_s$. In particular, packets sharing a public semantic readout need not be interchangeable at a recipient. This is an illustrative evaluation workflow, not an additional experiment. Appendix~\ref{app:formal} formalizes the conformance obligations, while Appendix~\ref{app:nip21-profile} presents a corresponding versioned compatibility-profile template.

\paragraph{What each check warrants.}
A public semantic-readout check provides evidence that evaluated packets satisfy the corresponding semantic requirement, subject to any additional contract-defined validity conditions. Natural-donor transplantation tests recipient continuation on the evaluated packets and permitted continuations. Both provide evidence limited to the tested conditions. Strong consumer conformance, by contrast, requires correct continuation for every protocol-valid packet and permitted continuation, as formalized in Appendix~\ref{app:formal}. Neither operational check alone establishes this universal obligation.

NIP does not mandate a particular neural architecture, private representation, exporter or importer design, supervision strategy, or training objective. Individual protocol versions may nevertheless fix packet schemas, semantic requirements, and exchange procedures that participating implementations must satisfy. Thus, implementations may differ in their private realization choices while remaining subject to the same public obligations at the designated computational boundary.

Participating implementations target the same task or computation and expose interfaces with a common protocol-defined computational role, while retaining flexibility in their private realizations. The experimental constructions use finite-dimensional packet spaces, deterministic exporters, identity transplantation, and semantic anchoring. These mechanisms characterize the studied realizations rather than universal requirements of NIP; individual protocols nevertheless impose their own specified packet and exchange obligations. The semantically indexed formal analysis additionally assumes that an intermediate semantic state, together with any permitted continuation input, suffices to determine the required downstream output. Appendix~\ref{app:formal} states this factorization and its assumptions precisely.

\subsection{Protocol Realization and Information-Access Rules}\label{app:nip-realization}
A published NIP can be implemented by distinct neural architectures while preserving a common packet-level exchange procedure. In the realization studied here, each donor exports a public packet from its private computation, and each recipient processes the unchanged packet through its own model-specific importer before downstream continuation. This exporter--importer decomposition specifies the evaluated implementation mechanism rather than a universal architectural requirement of NIP.

For donor implementation $i$ and recipient implementation $j$, let $h_i^{\mathrm{private}}$ denote the donor's intermediate state and $a(i)$ its protocol-declared state-shape class. The donor exports the public packet
\[
z_i^{\mathrm{public}}
=
E_{a(i)}\!\left(h_i^{\mathrm{private}}\right),
\]
which recipient $j$ processes through its own model-specific importer:
\[
\hat h_{i\rightarrow j}
=
I_j\!\left(z_i^{\mathrm{public}}\right).
\]
Here, $\hat h_{i\rightarrow j}$ is the recipient continuation state obtained from donor $i$'s packet, not a reconstruction of the donor's private representation.

\paragraph{Direct exchange through the public boundary.}
Cross-model transplantation replaces the recipient's public packet with the donor's naturally emitted packet, without a donor--recipient-specific transformation. The recipient applies its own importer and continues computation. This separates the shared exchange interface from implementation-specific representations and parameters.

Donor and recipient may differ in architecture, internal representation, or depth, provided they implement the same published packet requirements at boundaries with corresponding computational roles. In the evaluated protocols, transplantation is the identity operation in public packet space, followed by task-specific recipient continuation.

We use lowercase $z$ for vector-valued public packets and uppercase $Z$ for sequence-valued packets. Identity transplantation means exact substitution of the public packet, not substitution of arbitrary private hidden states. For implementations with different architectures or depths, the exchange boundary is identified by its designated computational role rather than necessarily by layer index.

\paragraph{Information access and development independence.}
Participating implementations may share the published protocol, permitted public task resources, and declared semantic supervision, including fixed exporters, packet schemas, or semantic readouts specified by the realization. Each implementation learns its own private parameters and importer. Development independence excludes reliance on other implementations' private representations and learned parameters; it does not require identity-disjoint public training datasets.

Under the independent-development condition, implementations do not use one another's checkpoints, private hidden states or activations, naturally emitted packets, learned importers, targets derived from another implementation's private representations, or donor--recipient-specific alignment losses. Post-hoc alignment baselines are evaluated separately under their stated information-access conditions. In the compact realization, the public exporter is determined by the declared shape class rather than fitted to incumbent features. During transplantation evaluation, model parameters and protocol requirements remain fixed, and donor-specific outcomes are not used to retune recipients.

In prospective studies, the public specification is fixed before the evaluated entrants are trained. This establishes a common development target, while successful compatibility remains subject to the specified qualification and continuation tests. Appendix~D documents the study-specific protocol freezes, exploratory investigations, and information exposure.

\paragraph{What is assessed at the interface.}
Producer-side checks evaluate whether naturally emitted packets satisfy specified public semantic requirements, without inspecting another implementation's private hidden states. Recipient-side tests assess downstream continuation after identity transplantation of independently emitted donor packets, using task-appropriate endpoints, including tasks with withheld continuation inputs. Matched corrupted-packet and mismatched-transfer controls provide evidence that task-relevant donor-packet information contributes to recipient performance under the tested interventions. These finite assessments do not establish universal producer or strong consumer conformance. In particular, packets with identical public semantic readouts may differ in decoder-invisible components that affect learned recipient behavior.

\paragraph{Training objectives and ablations in the tested realizations.}
NIP specifies interface obligations rather than a particular loss function. In the compact experiments, the objective in Eq.~\eqref{eq:compact-anchor} combines native-task supervision with a semantic-anchoring loss under a fixed public readout. Both terms use cross-entropy in this realization. Task supervision supports native computation, while the anchor term encourages naturally emitted packets to express the specified task semantics. The protocol-specified exporter remains fixed, and each model learns its own private parameters and importer. Neither successful optimization nor low semantic-readout loss establishes correct continuation for every packet in $\mathcal Z_s$.

The standard compact realization uses $\lambda_{\mathrm{anchor}}=1$. Setting $\lambda_{\mathrm{anchor}}=0$ removes the anchoring term while retaining task supervision, model capacity, bottleneck, and exporter. A separate intervention holds semantic anchoring fixed while changing the public exporter basis, testing sensitivity to the specified public coordinates. These ablations examine the contribution of particular realization mechanisms; they do not establish that semantic anchoring is required for every NIP implementation or exclude alternative compatibility mechanisms.

Structured-state tasks use task-specific semantic readouts. For continuous sequence-valued memories, a fixed semantic head is combined with permutation-invariant matching of producer-visible facts. Although these constraints differ mathematically, they provide public supervision for exchanged intermediate states while allowing implementation-specific private computation. Functional compatibility is evaluated through held-out donor substitution and task-specific controls, rather than inferred from the training losses alone.

\subsection{Illustrative Public Compatibility Profile}\label{app:nip21-profile}
\paragraph{Purpose and status.}
The illustrative NIP-4 compatibility profile, distinct from the experimentally evaluated protocols, demonstrates how a public neural interface contract could be expressed as a versioned specification and accompanying evaluation plan. Such a profile would publish the protocol identity, packet schema, computational boundary, semantic and validity requirements, permitted continuations, acceptance criteria, and reference testing procedures, giving independently developed implementations a common target.

Table~\ref{tab:nip21-profile} presents a proposed specification template, not an implemented or certified protocol. An operational profile would require concrete validity conditions, acceptance criteria, evaluation populations, and supported continuations to be declared when the protocol version is published. Finite admission and interoperability tests would provide evidence limited to their evaluation conditions, rather than establish universal producer or strong consumer conformance.

\begin{table}[tbp]
\centering
\caption{\textbf{Illustrative template for a four-dimensional NIP compatibility profile.} Proposed public specification fields and operational evaluation requirements, not an implemented standard or experimental result. Binding validity and acceptance criteria would be fixed when the protocol version is published.}
\label{tab:nip21-profile}
\scriptsize
\setlength{\tabcolsep}{3pt}
\begin{tabular}{p{0.30\linewidth}p{0.62\linewidth}}
\toprule
Profile element & Illustrative public requirement \\
\midrule
Identity and version & Unique protocol identifier and published version, including rules for identifying incompatible revisions. \\
Boundary and packet & Exchange boundary defined by computational role; four-dimensional packet schema specifying coordinate order, data type, and validity requirements. \\
Semantic interpretation & Public readout $D(z)$, semantic labels, and declared validity conditions, including requirements beyond decoder agreement where applicable. \\
Producer admission & Predeclared operational packet checks, evaluation inputs, and producer acceptance criteria. \\
Recipient continuation and acceptance & Supported continuation tasks, donor-packet substitution procedures, matched controls, evaluation endpoints, and acceptance criteria. \\
Evidence and limitations & Tested model and packet populations, uncertainty procedures, observed failures, and scope of the resulting compatibility claims. \\
\bottomrule
\end{tabular}
\end{table}

\paragraph{How an independent developer uses the profile.}
A specification publisher releases the versioned packet schema, computational boundary, semantic and validity requirements, supported continuations, acceptance criteria, and reference evaluation harness without distributing incumbent-private representations. A new producer is developed using permitted public resources and evaluated against the declared producer-admission checks. Recipients are then tested on their native packets and on naturally emitted packets from independently developed, admitted donors.

Evaluation outcomes are reported against the specified recipient endpoint, tolerance, protocol version, and tested population. Failed direct exchange is recorded rather than treated as successful interoperability after fitting a donor--recipient-specific transformation. Negative tests cover malformed packets, incompatible versions, and unsupported boundary roles. Changes to binding interface requirements are versioned and trigger evaluation against the revised contract rather than retrospective alteration of the original profile.

\paragraph{Typed application programming interface (API) versus neural packet contract.}
Conventional typed interfaces can specify packet structure, permissible interactions, and, where expressible, semantic invariants. A NIP additionally makes the functional use of exchanged intermediate information by independently trained neural continuations an explicit compatibility obligation. A four-dimensional packet with the required data type is therefore structurally well formed, but need not satisfy the contract's semantic requirements or support successful recipient continuation.

When the relevant intermediate state and continuation requirements can be specified completely in symbolic form, a conventional typed interface may offer a simpler and more readily verifiable solution. NIP is motivated by settings in which useful intermediate information is represented in learned states that are difficult to characterize exhaustively through symbolic rules. Neither the public contract nor the illustrative profile establishes correct consumer behavior on all unobserved protocol-valid packets.

\section{Additional Experimental Results}\label{app:step12-expanded-experiments}
\noindent\textbf{Guide to the experimental evidence.} This appendix follows four research questions: \emph{RQ1} examines direct exchange and prospective interoperability (Sections~\ref{app:c-direct-exchange}--\ref{app:c-prospective}); \emph{RQ2} examines withheld-information continuation and sequence-valued memory exchange (Sections~\ref{app:c-recipient-continuation}--\ref{app:c-sequence-memory}); \emph{RQ3} examines semantic sharing under partial public contracts (Section~\ref{app:c-contract-local}); and \emph{RQ4} examines decoder-invisible packet interventions (Section~\ref{app:c-behavioral-boundaries}).

Each study has distinct populations, controls, and endpoints; its results do not establish universal conformance. Subsections identify the principal EQs, although individual controls may inform multiple questions. Appendix~\ref{sec:supp-experiments} details experimental methods, and Appendix~\ref{app:formal} formalizes conformance.

\subsection{Direct Exchange and Its Mechanisms (RQ1; EQ2)}\label{app:c-direct-exchange}

\noindent\textbf{Identifying the mechanism of direct exchange.} We test whether independently trained models can exchange public packets without donor--recipient-specific alignment and which training mechanisms support that exchange. Two controls distinguish semantic anchoring from alternative explanations: removing the anchoring loss tests whether a shared compact bottleneck suffices, while changing the fixed public exporter basis tests sensitivity to coordinate geometry. The remaining task and model conditions are held fixed within each comparison.

\paragraph{Semantic anchoring supports direct exchange.}
Under the standard NIP-4 realization, models passing the native-task and same-model interface checks achieve probability-gap normalized recovery of $0.976$ with clean donor packets, versus $0.007$ with matched corrupted controls. The clean-versus-control advantage is $0.968$ (95\% CI $[0.957,0.979]$), with positive-edge density of $1.0$. NIP-8 and NIP-16 exhibit the same qualitative pattern. These results demonstrate effective direct exchange among the evaluated independently trained implementations under shared public semantic supervision.

\paragraph{A shared bottleneck alone does not ensure interoperability.}
Removing the semantic-anchoring loss while retaining the four-dimensional bottleneck, fixed exporter, task objective, and model capacity preserves native task competence and same-model recovery, but eliminates the observed cross-model advantage. Clean-donor probability-gap normalized recovery is $0.246$ versus $0.251$ for the corrupted control, yielding a clean-versus-control advantage of $-0.006$ (95\% CI $[-0.019,0.010]$); positive-edge density is $0.043$. Thus, a shared compact bottleneck and exporter are insufficient for direct exchange in this no-anchor realization.

\paragraph{Direct exchange is robust to a changed public basis.}
With semantic anchoring retained, replacing the canonical exporter with a fixed, publicly specified dense orthonormal projection preserves strong cross-model transfer. Clean-donor probability-gap normalized recovery is $0.995$ versus $0.0018$ for the corrupted control, yielding a clean-versus-control advantage of $0.993$ (95\% CI $[0.989,0.996]$); positive-edge density is $1.0$. The result shows that successful exchange in this realization does not depend on the canonical exporter coordinates.

Together, these ablations support semantic anchoring as an effective coordination mechanism in the tested realization: a shared compact bottleneck alone does not yield reliable cross-model exchange, while changing the fixed exporter basis preserves strong transfer when anchoring is retained. They do not establish that anchoring is necessary for all NIP implementations or that observed exchange proves universal conformance. Subsequent studies examine prospective compatibility with later-trained implementations and the extent of semantic sharing beyond explicitly constrained packet properties.

\paragraph{Post-hoc methods achieve strong transfer under different information-access conditions.}
A nonlinear pair-specific adapter and a four-dimensional shared post-hoc bus both achieve strong causal transfer (Table~\ref{tab:compact-mechanisms}), demonstrating that high recovery does not require specification-first training. However, these methods construct compatibility using incumbent representations or derived codes, which are unavailable during clean-room entrant development. NIP instead establishes a public contract before implementations are developed against it, enabling later models to target the interface without incumbent-private information. The comparison therefore distinguishes compatibility-construction settings rather than providing an information-matched performance ranking. It does not evaluate training-time compatible-representation methods based on predefined public heads or targets (see Section~\ref{sec:related}).

\begin{table}[t]
\centering
\scriptsize
\setlength{\tabcolsep}{4.5pt}
\renewcommand{\arraystretch}{1.15}
\caption{\textbf{Compatibility methods differ in their information-access and construction requirements.} The evaluated post-hoc methods use incumbent-derived representations, whereas NIP specifies a public interface before implementations are developed against it. Reported transfer advantages characterize performance under these distinct conditions; they are not an information-matched ranking or a comparison with all prospective compatible-representation methods.}

\label{tab:compact-mechanisms}
\begin{tabular}{lccccc}
\toprule
Method & \shortstack{Uses\\incumbent reps.?} & \shortstack{Pair-specific\\fitting?} & \shortstack{Prior public\\contract?} & \shortstack{Clean-room\\entrant?} & \shortstack{Transfer\\advantage} \\
\midrule
Nonlinear 4D adapter & Yes & Yes & No & No & $0.966$ \\
Shared post-hoc 4D bus & Yes & No & No & No & $0.952$ \\
NIP-4, canonical exporter & No & No & Yes & Yes & $0.968$ \\
NIP-4, dense exporter & No & No & Yes & Yes & $0.993$ \\
\bottomrule
\end{tabular}
\end{table}

\noindent\textbf{Finding and scope.} The controlled results support semantic anchoring as an effective mechanism for direct exchange, while the post-hoc baselines demonstrate alternative routes to compatibility under different information-access conditions. We next test whether later-trained implementations can use a previously fixed public protocol without access to incumbent-private representations.

\subsection{Prospective Interoperability (RQ1; EQ3, EQ5)}\label{app:c-prospective}

\noindent\textbf{Prospective exchange under a fixed specification.} We test whether later-trained implementations can exchange packets through an existing public interface without incumbent-private representations or donor--recipient-specific fitting. The experiments examine bidirectional ResNet--ConvNeXt exchange on CIFAR-100 and later-trained cohorts on dSprites. A separate identity-disjoint CIFAR-100 study evaluates recipient-native substitution using different models and endpoints; its results are interpreted independently of the prospective analytic-reference evaluation.

\paragraph{Prospective cross-architecture interoperability on natural images.}
On CIFAR-100, ten incumbent ResNets and ten later-trained ConvNeXts implement a shared 128-dimensional public packet with a fixed 64-dimensional semantic decoder. Images are processed before the exchange boundary, while continuation vectors are revealed afterward, requiring recipients to combine transplanted image states with newly available information. The ConvNeXt cohort is trained using public task resources and the fixed NIP specification, without incumbent-private representations or donor--recipient-specific alignment. Here, \emph{independently trained} refers to distinct implementations developed under a coordinated public task, protocol, and experimental recipe, not independently designed third-party systems. All 20 models pass native-task and same-model interface qualification, permitting evaluation of 100 cross-family donor--recipient pairs in each direction.

An exploratory development pair was inspected before the final evaluation plan was fixed. Subsequently, the 20-model population, 4,096 evaluation draws, 10,000-replicate model-node bootstrap, endpoints, acceptance criteria, and non-inferiority margin $\epsilon=0.03$ were frozen before confirmatory-cohort training or observation of cross-model outcomes. The margin came from an earlier within-model analytic-recovery tolerance, not the confirmatory results. We therefore describe an \emph{amended prospectively frozen confirmatory stage}, not an untouched preregistration. Some official CIFAR-100 test identities had also been encountered during earlier research and could recur across admission and final evaluation draws (Appendix~\ref{sec:supp-experiments}); this is not a pristine external test.

For each direction, we compare cross-model continuation with an analytic continuation derived from the donor's public semantic state:
\[
G_{\mathrm{analytic}} = R_{\mathrm{analytic}}-R_{\mathrm{cross}}.
\]
The frozen directional-mean non-inferiority criterion requires the upper 95\% model-node-bootstrap confidence bound for mean $G_{\mathrm{analytic}}$ to fall below $\epsilon=0.03$. Model-node resampling accounts for shared donors and recipients across the 100 substitutions per direction, but the intervals remain conditional on the selected evaluation images and do not guarantee coverage for unseen architecture families. The margin is an operational tolerance, not an externally validated minimum important difference.

Both directions meet the criterion: ResNet$\rightarrow$ConvNeXt has mean $G_{\mathrm{analytic}}=0.0212$ (95\% CI $[0.0197,0.0228]$), and ConvNeXt$\rightarrow$ResNet has $0.0186$ ($[0.0180,0.0194]$; Table~\ref{tab:e4-cifar}). This is a directional-mean result, not a per-pair guarantee; the maximum ResNet$\rightarrow$ConvNeXt gap is $0.03077$. Clean transfer also outperforms matched corrupted transfer, with causal advantages of $0.7019$ ($[0.6937,0.7110]$) and $0.9700$ ($[0.9578,0.9820]$), respectively, and positive-edge fractions of $0.98$ and $1.00$. Together, these results support direct, task-relevant exchange in the evaluated cohorts without establishing lossless substitution or universal conformance.

\paragraph{Keeping compatibility endpoints distinct.}
The analytic-reference gap, $G_{\mathrm{analytic}}=R_{\mathrm{analytic}}-R_{\mathrm{cross}}$, compares transferred continuation with a specified analytic reference. The recipient-native substitution gap, $g_{\mathrm{swap}}=R_{\mathrm{recipient\ clean}}-R_{\mathrm{transfer}}$, instead compares transferred performance with the recipient's own natural-packet continuation. Positive $g_{\mathrm{swap}}$ indicates degraded recipient-native recovery, while negative values indicate improvement. Because the reference conditions differ, meeting an analytic-reference tolerance does not guarantee recipient-native performance preservation. The separate identity-disjoint cross-architecture study uses $g_{\mathrm{swap}}$ as its primary directional endpoint; its results are not numerically pooled with $G_{\mathrm{analytic}}$. Neither metric establishes strong consumer conformance.

\begin{table}[t]
\centering
\scriptsize
\setlength{\tabcolsep}{1.8pt}
\renewcommand{\arraystretch}{1.08}
\caption{\textbf{Prospective CIFAR-100 cross-architecture exchange.}
The amended prospectively frozen evaluation includes ten models per architecture and 100 directed substitutions per direction. Both directional mean analytic-reference gaps meet the frozen non-inferiority criterion ($\epsilon=0.03$); intervals use model-node bootstrapping. Causal advantage compares clean with corrupted donor-packet transplantation. Positive-edge fractions refer to this contrast, not pair-level non-inferiority.}
\label{tab:e4-cifar}
\begin{tabular}{@{}l c c c@{}}
\toprule
Direction & $G_{\mathrm{analytic}}$ (95\% CI) & \shortstack{Causal advantage\\(95\% CI)} & \shortstack{Positive-edge\\fraction} \\
\midrule
Res.$\rightarrow$Conv. & $0.0212$ $[0.0197,0.0228]$ & $0.702$ $[0.694,0.711]$ & $0.98$ \\
Conv.$\rightarrow$Res. & $0.0186$ $[0.0180,0.0194]$ & $0.970$ $[0.958,0.982]$ & $1.00$ \\
\bottomrule
\end{tabular}
\end{table}

\paragraph{Later models can target the same public interface independently.}
On dSprites \citep{dsprites17}, we establish the public interface and incumbent population before training a later cohort from new random initializations using only public task resources and the published specification. Both cohorts satisfy the native-task and same-model interface criteria. Direct transplantation then achieves recovery of $0.9981$ for incumbent$\rightarrow$future, $0.9990$ for future$\rightarrow$incumbent, and $0.9986$ for future$\rightarrow$future exchange, with positive-edge density of $1.0$ in all three directions (Table~\ref{tab:e2c-cleanroom}). These results support prospective onboarding and bidirectional interoperability within the evaluated cohorts without incumbent-derived alignment.

\begin{table}[t]
\centering
\small
\setlength{\tabcolsep}{3.2pt}
\renewcommand{\arraystretch}{1.10}
\caption{\textbf{Prospective direct exchange on dSprites.}
The public interface is fixed before training the later cohort. Clean-donor and matched corrupted-control mean squared errors (MSEs) are reported for each exchange direction, alongside normalized squared-error recovery with 95\% model-node bootstrap confidence intervals.}
\label{tab:e2c-cleanroom}
\begin{tabular}{@{}l c c@{}}
\toprule
Direction (pairs) & \shortstack{Clean / control\\MSE} & \shortstack{Recovery\\(95\% CI)} \\
\midrule
Inc.$\rightarrow$Fut. (100) & $0.00119 / 0.1744$ & $0.9981$ $[0.9977,0.9984]$ \\
Fut.$\rightarrow$Inc. (100) & $0.00109 / 0.1739$ & $0.9990$ $[0.9988,0.9992]$ \\
Fut.$\rightarrow$Fut. (90) & $0.00110 / 0.1747$ & $0.9986$ $[0.9983,0.9988]$ \\
\bottomrule
\end{tabular}
\end{table}

\paragraph{Direct exchange extends across sequence architectures.}
We replace the private Transformer with an independently trained two-layer GRU while retaining the public interface. After both model families pass native-task and same-model interface qualification, Transformer$\rightarrow$GRU exchange achieves a causal advantage of $0.832$ (95\% CI $[0.802,0.862]$), and GRU$\rightarrow$Transformer achieves $0.790$ ($[0.768,0.813]$); all evaluated edges are positive in both directions. These results demonstrate successful direct exchange across the tested sequence architectures. Additional depth and width variations and a computational-role boundary-placement control are reported in Appendix~\ref{app:implementation-variation}.

\noindent\textbf{Finding and scope.} The evaluated cohorts support prospective interoperability and direct exchange across selected architectures under a fixed public interface, without establishing compatibility for arbitrary implementations. We next test whether recipients can continue computation using packets emitted before donors receive the information needed to determine the final answer.

\subsection{Recipient-Side Continuation (RQ2; EQ8)}\label{app:c-recipient-continuation}

\paragraph{Why nonterminal continuation is a distinct test.}
We test whether a recipient can perform further computation using another model's intermediate packet, rather than merely reuse features for a terminal prediction. The donor emits its packet before the continuation input needed to determine the final answer becomes available; the recipient then combines that packet with the subsequently revealed input. This tests \emph{nonterminal interoperability} without implying that other compatible-representation methods could not support the same capability.

\paragraph{The donor cannot know the realized answer at the interface.}
The task has two stages,
\[
\mathrm{source}\xrightarrow{r_1}\mathrm{intermediate}\xrightarrow{r_2}\mathrm{final},
\]
where the donor observes the source and $r_1$, but not the later continuation $r_2$. For any fixed donor-visible state, the four possible values of $r_2$ yield four equiprobable final answers. Thus, a predictor restricted to donor-visible information has maximum accuracy $0.25$. This rules out a donor-only final-answer shortcut while allowing the packet to encode a sufficient intermediate state for recipient continuation.

\paragraph{Transferred packets support nonterminal computation across models.}
All models pass native-task and same-model interface qualification, with same-model recovery of $1.0$. Across 2,450 ordered donor--recipient pairs, clean-donor probability-gap normalized recovery is $0.838$ versus $0.049$ for matched corrupted controls, yielding a clean-versus-control advantage of $0.789$ (95\% CI $[0.732,0.846]$). Because donors never observe $r_2$, successful transplantation provides evidence that recipients combine transferred intermediate information with the subsequently revealed continuation input. This establishes task-relevant nonterminal exchange for the evaluated implementations and naturally emitted packets, not compatibility with arbitrary intermediate states or decoder-consistent packets.

\paragraph{Nonterminal exchange does not require dedicated factor coordinates.}
A structured 16-valued variant and a separate dense-decoder realization reproduce the qualitative continuation result, including without an explicit assignment of packet coordinates to individual semantic factors. These controls show that the observed effect is not restricted to the original four-way output or a hand-assigned factor layout; both realizations retain task-specific public semantic supervision. Supporting results appear in Appendix~\ref{app:structured-onboarding}.

\noindent\textbf{Finding and scope.} The withheld-continuation design excludes transmission of a donor-known final answer, while allowing the packet to carry a sufficient intermediate state for recipient computation. We next examine whether this form of nonterminal exchange extends from compact packets to sequence-valued continuous memory.

\subsection{Sequence-Valued Continuous Memory (RQ2; EQ9)}\label{app:c-sequence-memory}

\noindent\textbf{Continuation through sequence-valued memory.} We extend nonterminal exchange from compact packets to a sequence-valued continuous-memory interface. Each example contains four key--value facts with four-dimensional continuous values, represented by a public four-token packet $Z\in\mathbb{R}^{4\times16}$. The producer observes the facts but not the subsequent query or continuation, requiring the recipient to use the transferred memory to compute the requested output. We evaluate this interface prospectively under a fixed protocol and analysis plan; Appendix~\ref{app:continuous-details} details the task, training losses, and evaluation endpoints.

\paragraph{Continuous memories transfer bidirectionally across independently trained models.}
For incumbent pairs, clean transplantation yields MSE of $0.0085$ versus $0.2995$ under matched corruption, giving a corrupted-minus-clean MSE advantage of $0.291$ (95\% CI $[0.286,0.296]$). The later clean-room cohort shows comparable advantages in both directions: $0.291$ ($[0.287,0.296]$) for future$\rightarrow$incumbent and $0.291$ ($[0.286,0.295]$) for incumbent$\rightarrow$future, with positive contrasts on every evaluated edge. These results support prospective exchange of naturally emitted sequence-valued memories in the tested setting, without incumbent-private representation access.

\noindent\textbf{Finding and scope.} The evaluated models support prospective exchange of naturally emitted sequence-valued memories, without establishing that the public readout captures all information used by recipients. We next examine whether interoperability extends beyond the semantics explicitly specified by the contract.

\subsection{Contract-Local Semantic Interoperability (RQ3; EQ11)}\label{app:c-contract-local}

The preceding experiments demonstrate direct, prospective, and nonterminal exchange in the evaluated settings. We now ask whether interoperability extends beyond the semantics explicitly specified by the contract. A partial specification might indirectly align other task-relevant information, or exchange might remain \textbf{contract-local}, with unspecified factors retaining model-private representations. We test this distinction empirically rather than assuming that unconstrained factors must be incompatible.

\paragraph{Controlled latent-factor task.}
We use four independent categorical factors $s=(s_1,s_2,s_3,s_4)$, each with four possible values. After packet emission, the recipient receives a query $q\in\{1,2,3,4\}$ and must predict $y=s_q$. Because the producer does not know which factor will be queried, successful same-model continuation requires retaining information about all four factors. We hold out latent-factor combinations during training and require strong held-out task performance and same-model packet sufficiency before testing cross-model exchange.

\paragraph{Partial public contracts.}
We vary how many task-relevant factors the public contract explicitly constrains, holding packet dimensionality, task structure, and evaluation fixed. The $0\%$, $25\%$, $50\%$, $75\%$, and $100\%$ conditions anchor zero through four factors, respectively. Anchors are fixed public readouts, not learned aligners or targets derived from incumbent representations. Cross-model compatibility for an unanchored factor would therefore arise without an explicit semantic constraint on that factor.

\paragraph{Separating information preservation from shared geometry.}
For populations passing native-task and same-model interface qualification, we evaluate direct transplantation across the full $4^4=256$ factorial state space, separately measuring anchored and unanchored query accuracy, with anchor-matched shuffled-donor and counterfactual controls. Within-model linear probes test whether each factor remains recoverable from its producer's packets; probes trained on one model and applied unchanged to another test cross-model decodability. Together with direct recipient continuation, these distinguish \emph{information retention}, \emph{cross-model decodability}, and \emph{operational interchangeability}. Recoverability within a model alone guarantees neither of the latter properties. This experiment concerns factors omitted from the public contract, not arbitrary packet variations that preserve its specified semantic readout.

\paragraph{Independent replication and nonlinear stress test.}
After exploratory identification of the anchor-fraction effect, we tested it in new independent model populations using a fixed training and evaluation procedure. We varied both the anchored factor identities and their physical input positions to assess dependence on particular factors or locations. A separate robustness test applied deterministic nonlinear cross-coordinate mixing at the packet boundary while preserving the task and public semantic obligations.

\begin{figure}[!t]
\centering
\includegraphics[width=0.94\linewidth]{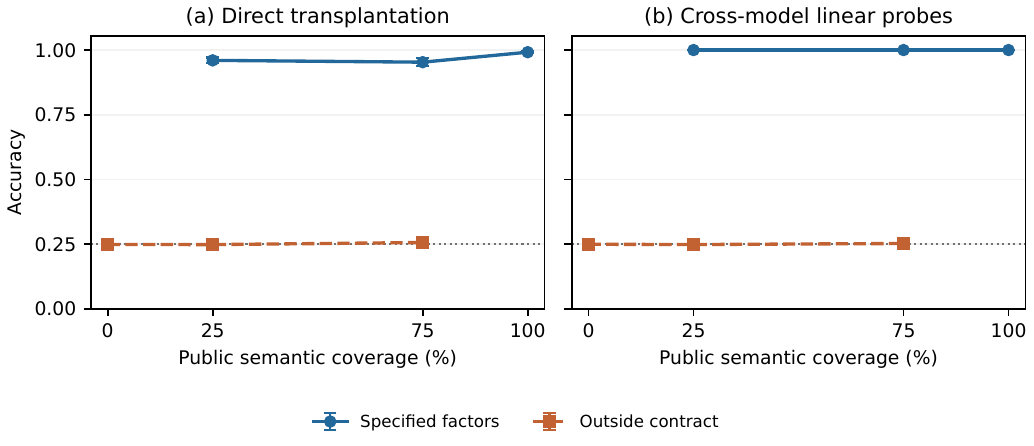}
\captionof{figure}{\textbf{Interoperability is local to the public contract.}
Across four independent replication populations with rotated anchor identities and physical input positions, explicitly specified factors exhibit high direct-transplantation and cross-model-probe accuracy, while unspecified factors remain near four-way chance (25\%). Markers report means across four independent populations; error bars are descriptive 95\% Student-$t$ confidence intervals over population-level estimates ($n=4$, 3 degrees of freedom), clipped to the accuracy range $[0,1]$. Donor--recipient pairs are not treated as independent replicates. The dotted line marks four-way chance (25\%). The replicated coverage conditions are $0\%$, $25\%$, $75\%$, and $100\%$; absent endpoint series are undefined because the corresponding factor category is absent. At $75\%$ coverage, specified-factor transplantation is $0.9532$ versus $0.2563$ for the omitted factor; cross-model probing is $1.000$ versus $0.2525$. Within-model probes recover all factors essentially perfectly, while a separate nonlinear packet-mixing test retains the qualitative separation.}
\label{fig:contract-locality}
\end{figure}

\paragraph{Interoperability is contract-local.}
Figure~\ref{fig:contract-locality} summarizes four independent replication populations. Unanchored factors remain near four-way chance: direct-transfer accuracy is $0.2483$ at $0\%$ anchoring, $0.2483$ at $25\%$, and $0.2563$ at $75\%$. Anchored-factor accuracy is substantially higher: $0.9602$, $0.9532$, and $0.9918$ at $25\%$, $75\%$, and $100\%$ coverage, respectively. Unchanged cross-model probes show the same separation, achieving $1.000$ accuracy on anchored factors but only $0.2495$, $0.2484$, and $0.2525$ on unanchored factors. Thus, specifying three of four equally task-relevant factors does not necessarily make the remaining factor interoperable, even when each model retains that information.

The replication includes four independent populations. In the $0\%$ condition, 38/40 models pass native-task and same-model interface qualification; all models pass in the remaining conditions. Relative to $0\%$ anchoring, changes in unanchored direct-transfer accuracy are $-0.0001\pm0.0098$ at $25\%$ and $0.0079\pm0.0102$ at $75\%$ (mean $\pm$ population SD), showing no consistent spillover in these populations. The qualitative separation persists under deterministic nonlinear packet mixing: anchored-factor transfer remains high ($0.9200$ at $25\%$ and $0.9172$ at $75\%$), while unanchored transfer and unchanged cross-model probes remain near chance. Together with successful within-model factor recovery, these findings distinguish \emph{information preservation} from \emph{operational interoperability} under the tested partial contracts.

\noindent\textbf{Finding and scope.} The evaluated partial contracts preserve unanchored task information within individual models without making it reliably interoperable across implementations. We next test a distinct boundary: whether recipients behave identically on packets with the same publicly decoded semantics but different decoder-invisible components.

\subsection{Behavioral Conformance Boundaries (RQ4; EQ13)}\label{app:c-behavioral-boundaries}

We next distinguish two forms of under-specification. Appendix C.5 tests task-relevant factors omitted from a partial public contract; here, we examine packet components invisible to a specified public semantic decoder. In neither case does the absence of an explicit constraint alone determine whether recipient behavior changes.

\paragraph{Naturally emitted decoder-invisible variation can be benign.}
We replace each continuous-memory token's decoder-null component with one naturally emitted by another donor and matched to the same semantic fact. Clean-transfer MSE changes from $0.008485$ to $0.008427$, a swap-minus-original difference of $-5.8\times10^{-5}$ (95\% CI $[-6.4\times10^{-4},5.1\times10^{-4}]$); decoded semantics change by at most $5.7\times10^{-6}$. The tested recipients show no detectable deterioration under these matched natural swaps, supporting tolerance of some decoder-invisible variation without establishing invariance to arbitrary null-component changes.

\paragraph{Decoder agreement does not suffice for arbitrary interventions.}
Two synthetic interventions on continuous-memory packets increase clean-transfer MSE: projection onto the public decoder row space raises it to $0.0153$, and matched-norm randomization of decoder-null components raises it to $0.0266$, despite decoded-semantic changes below $9\times10^{-6}$. Both retain positive clean-versus-corrupted contrasts, indicating increased prediction error rather than complete loss of usable transferred information. Unlike matched natural swaps, these interventions may leave the producer-packet distribution. Thus, agreement under the public semantic decoder does not guarantee equal recipient behavior, nor does it establish that the modified packets satisfy every protocol-validity obligation.

\paragraph{Two boundaries, one interpretation.}
Partial-contract experiments show that task-relevant information can remain recoverable within models without becoming interoperable across them. Continuous-memory interventions show that naturally emitted decoder-invisible variation can be tolerated, while synthetic variation can affect recipient behavior despite near-identical public semantic readouts. These results distinguish \emph{specified semantics}, \emph{within-model information retention}, and \emph{recipient behavior} without establishing either universal incompatibility of omitted factors or strong conformance over the protocol-valid set $\mathcal Z_s$. Decoder agreement alone does not establish membership in $\mathcal Z_s$, which is defined by all public protocol obligations. Appendix~\ref{app:formal} states the conditions for strong conformance; Appendix~\ref{app:continuous-controls} details the packet interventions.

\section{Experimental Protocols and Supporting Analyses}
\label{sec:supp-experiments}

\textbf{How to read the experimental protocols.} This appendix documents the evaluation designs, information-access boundaries, controls, and supporting analyses for each research question.

\paragraph{Units of analysis and limits of inference.}
Cross-model exchanges reuse donors, recipients, and evaluation examples. A directed donor--recipient edge is therefore a tested substitution, not an independent model-training replicate. Reported confidence intervals use model-node, hierarchical, or study-specific resampling to account for relevant dependencies, subject to each study's model selection, evaluation population, and resampling assumptions.

The amended prospective 20-model CIFAR-100 study, separate eight-model identity-disjoint study, and exploratory training-population-matched analyses differ in design and endpoints; they do not estimate a single common effect. Later-cohort onboarding likewise tests separately trained implementations under a shared public protocol, not independently commissioned third-party systems.

\paragraph{Research-question guide.}
Appendix D provides supporting protocols and analyses for RQ1 (direct and prospective exchange), RQ2 (recipient-side continuation), RQ3 (semantic and computational boundaries), and RQ4 (scope of conformance evidence). The evaluation design and information-access rules appear first (Appendices~\ref{app:frozen-controls}--\ref{app:nip21-information}), followed by experiment-specific methods and controls. Appendix~\ref{app:formal} presents the formal conformance conditions.

\subsection{Evaluation Design and Reproducibility}
\label{app:frozen-controls}
\label{app:reproducibility}

\textbf{Interpreting the experimental designs.} Controlled ablations investigate mechanisms, exploratory studies compare design choices, and prospectively specified studies evaluate fixed protocols and endpoints in later model populations. These categories reflect each study's documented development and evaluation chronology, not an assertion that the research program was free of earlier exploratory exposure. Table~\ref{tab:confirmatory-design} summarizes the study designs.

\textbf{Confirmatory analyses are separated from development and post-hoc fitting.} Evaluation plans are frozen before their designated confirmatory tests, but not necessarily before model training or architecture selection. For the amended prospective CIFAR-100 study, the final plan followed an excluded development-pair inspection but preceded confirmatory training. In structured retrieval and identity-disjoint CIFAR-100, model populations and held-out procedures were fixed after development and before sealed evaluation. Cross-model tests generally follow native-task and same-model interface qualification; EQ6 is an exception, retaining all 50 two-layer GRUs despite two failing individual behavioral qualification. NIP evaluation excludes donor--recipient-specific fitting, while post-hoc baselines use disjoint fitting and evaluation cases.

Model-node bootstrapping accounts for dependence among donor--recipient exchanges that share trained models. Reported confidence intervals remain conditional on the evaluated populations and test examples.

\begin{table}[tbp]
\centering
\caption{\textbf{Study design and evaluation chronology.} Representative study populations, endpoints, and evaluation histories. Mechanism-discovery analyses and post-hoc baselines are reported separately.}
\label{tab:confirmatory-design}
\scriptsize
\setlength{\tabcolsep}{3pt}
\begin{tabular}{@{}p{0.15\linewidth}p{0.19\linewidth}p{0.24\linewidth}p{0.33\linewidth}@{}}
\toprule
Study & Evaluated population & Primary evidence & Evaluation chronology \\
\midrule
dSprites & Incumbent and later cohorts & Bidirectional transplantation & Protocol and clean-room evaluation frozen before confirmatory cohorts \\
CIFAR-100 & 10 ResNets; 10 later ConvNeXts & Analytic-reference gap; model-node bootstrap & Development pair excluded; amended plan frozen before confirmatory training \\
Continuous memory & 30 incumbent; 30 later models & Direct exchange; model-node bootstrap & Protocol and primary endpoint frozen before training \\
Nonterminal continuation & Evaluated directed producer--recipient pairs & Withheld-information continuation & Controlled test; interval resampling requires record-level verification \\
Structured retrieval & Eight admitted models; two objectives & Held-out direct exchange & Family selected after development; source-disjoint evaluation after admission \\
\bottomrule
\end{tabular}
\end{table}

\textbf{dSprites required an identifiable semantic target.} Rotationally symmetric shapes can produce identical images at different simulator orientations, making raw orientation unidentifiable from the image. The public orientation state is therefore defined modulo each shape's rotational symmetry. This definition was fixed before confirmatory clean-room training, and the later cohort used only the public task and protocol specification.

\textbf{CIFAR-100 admission and evaluation data provenance.} Confirmatory models are trained on the CIFAR-100 training split. Pre-transfer admission uses 2,048 draws from the official test split per model (seeds 20400302 for ResNet and 20400502 for ConvNeXt); final cross-model evaluation uses a separate 4,096-draw sample from the same split (seed 20400701). Sampling is with replacement, so images may recur within and across samples despite different seeds. Although admission does not use cross-model transfer outcomes, it does use official test-split images. The confirmatory findings are therefore conditional on this prespecified admission procedure, not on a test split reserved exclusively for final evaluation.

\textbf{CIFAR-100 uses an amended prospectively frozen confirmatory stage.} An inspected ResNet/ConvNeXt development pair was excluded before the confirmatory plan was frozen. The fixed populations comprise ResNet seeds 8800--8809 and ConvNeXt seeds 8900--8909, without replacements; all 20 models passed prespecified pre-transfer qualification. Directional-mean non-inferiority requires the upper 95\% model-node-bootstrap confidence bound for $G_{\mathrm{analytic}}$ to fall below $\epsilon=0.03$. This margin came from an earlier within-model analytic-recovery tolerance, not confirmatory outcomes, and is an operational rather than externally validated threshold. The causal endpoint requires the lower confidence bound for clean-minus-corrupted transfer to exceed zero.

\textbf{The contract-locality experiments address a different question.} An exploratory partial-anchor effect was replicated in independent populations at 0\%, 25\%, 75\%, and 100\% anchoring, with factor identities and input positions rotated. A separate nonlinear-mixing test examined robustness to packet-coordinate changes. These results show that unanchored information can remain recoverable within models without becoming interoperable across them; they neither establish universal incompatibility nor substitute for prospective clean-room evaluation.

\paragraph{Relationship between experimental endpoints and uncertainty.}
The prospective CIFAR-100 study evaluates an analytic-reference gap across 200 directed substitutions using model-node bootstrap intervals. The separate training-population-matched analysis measures recipient-native substitution on previously examined development identities; its image-bootstrap intervals condition on fixed checkpoints and exclude between-seed variability. These studies differ in populations, evaluation histories, and endpoints, so their results are not independent replications of one effect. The earlier shared-classifier classification readout is likewise not a matched method-level comparator.
\subsection{Information Access and Selection Boundaries}\label{app:nip21-information}
\paragraph{Comparing interface-construction information.}
The evaluated NIP implementations use a fixed public interface and semantic supervision without incumbent-private representations. In contrast, the post-hoc adapters and incumbent-fitted shared buses in Table~\ref{tab:compact-mechanisms} use existing-model representations during compatibility construction. Prospective fixed-head methods could also use comparable public supervision, but no information- and supervision-matched implementation was evaluated. These comparisons therefore distinguish information-access conditions, not general method superiority.
\paragraph{Method selection and evaluation separation.}
In structured retrieval, a query-conditioned architecture replaced a generic architecture that failed admission; eight admitted models were then frozen before source-disjoint evaluation. The 7,168 successful pair--source outcomes reuse eight models and 128 semantic sources, so the saturated bootstrap interval is conditional on that selected population. The identity-disjoint CIFAR-100 study trained a fresh eight-model cohort using official training images and admission-restricted diagnostics, but inherited architecture and recipe choices from earlier development involving official test images. Its pre-evaluation commitment prevents subsequent selection on final outcomes, not prior research-program exposure. Three-axis bootstrap intervals account for shared donors, recipients, and queries but remain descriptive given four models per architecture and replacement-sampled queries.
\paragraph{Endpoints are not interchangeable.}
Structured retrieval reports exact-answer accuracy, whereas identity-disjoint CIFAR-100 reports normalized squared-error recovery, which need not lie in $[0,1]$. The identity-disjoint study's recipient-native substitution gap also differs from the analytic-reference gap used in the 20-model prospective CIFAR-100 study. The latter's $0.03$ non-inferiority margin and model-node bootstrap inference therefore do not apply to the identity-disjoint evaluation.
\subsection{Matched-Task-Data Structured-Retrieval Control (EQ1)}\label{app:matched-information}
\paragraph{Matched task information with different semantic objectives.}
We test whether explicit public semantic-head supervision improves direct exchange beyond native-task learning alone. Task-only and semantic-head models use the same query-conditioned architecture (2,816 parameters), four seeds (701--704), 360 training sources, 108 admission sources, and 2,500 updates per model with batch size 128. A separate set of 108 semantic sources is reserved for held-out evaluation and used for neither training nor admission.

Each instance contains two distinct key--value bindings and a query selecting one of eight keys, with four possible value classes. The public packet $z\in\mathbb{R}^4$ contains continuous logits; the semantic-head arm adds $\operatorname{CE}(z,y)$, for queried value $y$, to the native classification loss, fixing the public coordinate-to-class interpretation. Both arms use the same task examples and labels, but differ in intermediate semantic supervision and training loss. This tests one semantic-anchoring realization without pair-specific alignment or recipient fine-tuning, not superiority over all prospective compatible-feature methods.

For the corrupted-donor control, only the queried binding's value is shifted to the next class modulo four before packet generation; predictions remain scored against the original target.

Table~\ref{tab:matched-information} summarizes held-out transfer across the task-only and semantic-head donor--recipient combinations. The prespecified primary endpoint compares mean direct exact-answer accuracy across 12 distinct directed within-arm pairs per arm, semantic-head minus task-only. Eight diagonal self-pairs serve as controls; all 64 ordered pairs, including 32 mixed-arm pairs, are retained for secondary analysis. A 2,000-replicate hierarchical bootstrap accounts for shared sources and model identities, with descriptive intervals given only four models per arm. Model populations, the primary endpoint, and the analysis procedure were fixed before held-out evaluation.

\begin{table}[t]
\centering
\caption{\textbf{Matched-task-data held-out transfer.} Mean exact-answer accuracy over 108 held-out semantic source identities for off-diagonal donor--recipient pairs. Corrupted controls use the same pair assignments, and all eight models achieve 100\% native-task accuracy.}
\label{tab:matched-information}
\scriptsize
\setlength{\tabcolsep}{3pt}
\begin{tabular}{@{}p{0.40\linewidth}rrr@{}}
\toprule
Donor $\to$ recipient & Pairs & Transfer & Corrupted \\
\midrule
Task-only $\to$ task-only & 12 & $35.65\%$ & $25.00\%$ \\
Task-only $\to$ semantic-head & 16 & $21.06\%$ & $12.44\%$ \\
Semantic-head $\to$ task-only & 16 & $15.39\%$ & $29.69\%$ \\
Semantic-head $\to$ semantic-head & 12 & $100.00\%$ & $0.00\%$ \\
\bottomrule
\end{tabular}
\end{table}

\paragraph{Results and interpretation.}
On 108 unique held-out sources, within-arm direct-transfer accuracy is $1.000$ for semantic-head models versus $0.3565$ for task-only models, a difference of $0.6435$ (95\% descriptive bootstrap interval $[0.2778,1.0]$); the qualification-set difference is $0.6366$. All models achieve perfect native-task accuracy, but public semantic accuracy is $100\%$ for semantic-head models versus approximately $14.35\%$ for task-only models (individual rates 0, $28.70\%$, 0, $28.70\%$). Task-only models exhibit seed-dependent packet semantics and incompatible cross-model continuations. Corrupted-donor accuracy is $0\%$ within semantic-head pairs, while other blocks include nonzero corrupted accuracy and negative clean-versus-corrupted contrasts. Thus, explicit semantic supervision supports strong within-arm interoperability in this setting, but even a publicly interpretable packet need not work with a recipient trained under a different private convention.

\paragraph{Scope and reproducibility.}
Held-out evaluation covers 64 ordered pairs, including 56 off-diagonal pairs, across 108 distinct semantic sources. Checkpoints, pair identities, the primary contrast, and bootstrap seed 2117026 were fixed before evaluation, with per-source correctness records retained for verification. The findings remain specific to the selected model family and source partition, with uncertainty limited by four independently trained models per arm.
\subsection{Additional Compact-Interface Evidence}
\label{app:compact-evidence}

\textbf{The mechanism controls separate compression, sufficiency, and interoperability.} Fixed 2D, 4D, and 8D post-hoc hidden-state projections fail same-model sufficiency, although the full 16D state remains useful; training through the interface instead yields usable compact packets at all three tested dimensions. No-anchor models retain native-task competence and same-model restoration but lose cross-model transfer, which returns with shared semantic anchoring and a dense exporter. These controls support the role of a common semantic convention in this realization, without ruling out other compression methods or prospective compatibility mechanisms.

\textbf{Learned post-hoc mappings also achieve strong causal transfer.} With disjoint fitting and evaluation data, a rank-4 affine adapter achieves a clean-minus-control advantage of $0.868$ (95\% CI $[0.836,0.899]$), while a 356-parameter nonlinear 4D-bottleneck MLP achieves $0.966$ ($[0.953,0.978]$); both have positive-edge density $1.0$. These pair-specific baselines demonstrate strong transfer without prospective protocol training. The distinction from NIP is how compatibility is established, not whether effective transfer is possible.

\textbf{A reusable post-hoc shared bus provides a non-pairwise baseline.} A shared 4D space learned from incumbent training-fold activations achieves an incumbent transfer advantage of $0.952$ (95\% CI $[0.935,0.967]$), with positive-edge density $1.0$. Once fixed, the bus supports later models through fitted model-specific adapters, achieving future$\to$incumbent advantage $0.935$ ($[0.917,0.952]$) and incumbent$\to$future advantage $0.957$ ($[0.943,0.969]$). Fitting and evaluation cases are disjoint. Unlike NIP's independently specified public interface, this reusable compatibility target is derived from incumbent representations.
\subsection{Training-Population-Matched NIP Development Analysis}
\label{app:matched-split-nip}

\paragraph{Design and inferential role.}
Earlier exploratory NIP models used all 50,000 CIFAR-100 training images, including 5,000 identities held out from shared-classifier diagnostic training. To remove this training-population mismatch, we trained a new ResNet-18 (seed 9500; 20,000 updates) and ConvNeXt (seed 9600; 30,000 updates) on the same 45,000 fitting identities, reserving the remaining 5,000 for development evaluation. Both follow the original family-specific recipes and unchanged public NIP protocol, including its fixed $64\times128$ semantic decoder, without pair-specific alignment, cross-model feedback, or recipient adaptation. This exploratory analysis uses only official training-split images and does not access the official test split.

\paragraph{Measurements.}
Table~\ref{tab:matched-split-nip} reports the native controls and directed substitutions in this training-population-matched exploratory analysis. We evaluate both native controls and directed substitutions on the same development images, using four normalized 64-dimensional continuation queries per image (seed 20310106). The target is the query's dot product with eight times the public class codebook. Writing $L_{d\to r}$ for mean squared continuation error, donor-relative excess MSE is $L_{d\to r}-L_{d\to d}$, while recipient-relative excess MSE is $L_{d\to r}-L_{r\to r}$. Recovery is normalized against zero-predictor MSE $1.0016$. Percentile confidence intervals use 5,000 image-bootstrap replicates (seed 20310105), resampling each image with its four queries; they condition on the two fixed model checkpoints.

\begin{table}[t]
\centering
\caption{\textbf{Training-population-matched exploratory NIP transfer.} All results use 5,000 development images excluded from both models' training sets. Donor excess MSE compares cross-model continuation with the donor's native continuation. Confidence intervals use image bootstrap and are conditional on the two fixed model checkpoints.}
\label{tab:matched-split-nip}
\scriptsize
\setlength{\tabcolsep}{3pt}
\begin{tabular}{@{}lrr@{}}
\toprule
Producer$\to$recipient & Recovery & Donor excess MSE (95\% CI) \\
\midrule
ResNet$\to$ResNet & $0.3507$ & --- \\
ResNet$\to$ConvNeXt & $0.3389$ & $0.0119$ $[0.0066,0.0172]$ \\
ConvNeXt$\to$ConvNeXt & $0.4901$ & --- \\
ConvNeXt$\to$ResNet & $0.4750$ & $0.0151$ $[0.0127,0.0175]$ \\
\bottomrule
\end{tabular}
\end{table}

Both directional upper confidence bounds fall below an exploratory $0.05$ tolerance, distinct from EQ3's frozen $0.03$ analytic-reference criterion. Recipient-relative excess MSE is $+0.1515$ for ResNet$\to$ConvNeXt and $-0.1245$ for ConvNeXt$\to$ResNet, reflecting differences in native model performance. Transferred predictions also differ from recipient-native outputs (disagreement MSE $0.2536$ and $0.2652$, respectively), so preserved recovery does not imply output equivalence. Given prior exposure to the development partition and inherited training recipes, these findings constitute exploratory held-out-from-training evidence, not a fresh confirmatory replication.

\subsection{Structured Retrieval and CIFAR-100 Evaluation Protocols (EQ4, EQ7, EQ12)}
\label{app:m20-methods}

\paragraph{Structured associative retrieval across training objectives.}
The task uses a public four-dimensional packet representing a binding over five keys and four values, emitted before recipient continuation. Its 640 semantic source identities are partitioned into 384 training, 128 admission, and 128 held-out sources, with disjoint identities across partitions. Four independently initialized models use fixed-head cross-entropy and four use a calibrated Brier objective under the same public protocol. The query-conditioned architecture was selected during development after a generic Transformer failed unseen-source admission; all eight selected models subsequently passed source-disjoint qualification. Before held-out evaluation, the model population and transfer rules were frozen. We test 56 directed off-diagonal pairs (24 within-objective and 32 cross-objective), plus eight diagonal controls, using identity transplantation without fitted mappings and comparing clean transfer with wrong-target donor and canonical semantic packets. The study tests objective variation within the selected architecture family, not arbitrary architecture independence.

\paragraph{Architecture substitution with identity-disjoint CIFAR-100 data.}
Separately from the 20-model prospective CIFAR-100 study, we train four ResNet-18 and four ConvNeXt models (seeds 301--304 per family) on the official training split, using the previously developed ABI-128 protocol, a fixed 64-dimensional public semantic readout, and family-specific recipes. This study uses different endpoints and no inherited non-inferiority margin. Because the recipes and architecture choices derive partly from earlier development involving official test images, the evaluation is not fully independent of the research program's prior test exposure.

We partition the official 10,000-image test split into 5,000 admission and 5,000 held-out identities, with 50 examples per class in each and exact raw-pixel duplicate groups kept together, including groups with conflicting labels. All eight models pass admission without pair-specific adaptation. After inspecting cross-architecture transfer on admission data, we freeze the complete checkpoint population and final evaluation rules. Held-out evaluation covers all 64 ordered donor--recipient pairs (56 off-diagonal, including 32 cross-architecture) using 5,000 shared queries sampled with replacement from the eligible held-out identities; the draws need not represent 5,000 distinct images.

\paragraph{Prespecified recovery endpoints and analysis.}
For each evaluation condition $k$, let $E_k=\sum_q(\hat y_{k,q}-y_q)^2$ denote summed squared prediction error over the shared queries. When $\sum_q y_q^2>0$, normalized squared-error recovery is
\begin{equation}
R_k=1-\frac{E_k}{\sum_{q} y_q^2},
\label{eq:m20b-recovery}
\end{equation}
This continuous-task measure is neither CIFAR-100 classification accuracy nor the probability-gap recovery defined in Eq.~\eqref{eq:recovery}. The primary recipient-native substitution gap is
\begin{equation}
g_{\mathrm{swap}}=R_{\mathrm{recipient\,native}}-R_{\mathrm{transfer}},
\label{eq:m20b-swap}
\end{equation}
where positive values indicate worse recovery after donor-packet substitution and negative values indicate improvement. Companion endpoints are $g_{\mathrm{analytic}}=R_{\mathrm{analytic}}-R_{\mathrm{transfer}}$ and the clean-versus-corrupted contrast $R_{\mathrm{transfer}}-R_{\mathrm{corrupt}}$. The two cross-architecture directions are analyzed separately. Neither the analytic-reference gap $G_{\mathrm{analytic}}$ nor the frozen $\epsilon=0.03$ non-inferiority margin from the earlier prospective CIFAR-100 study applies here. Uncertainty summaries use 1,000 bootstrap replicates across donor, recipient, and shared-query axes; intervals remain descriptive given four models per architecture and repeated queries.

\paragraph{Prespecification and evaluation integrity.}
Model populations, identity partitions, and analysis rules were fixed before held-out evaluation. Following a serialization failure, aggregate statistics were reconstructed from saved per-query error arrays using the frozen analysis procedure, without further held-out image access or model inference.

\subsection{Statistical Analysis and Study Limitations}\label{app:m20-audit}

\paragraph{Structured-retrieval evaluation and dependence.}
Eight admitted models yield $8\times7\times128=7{,}168$ off-diagonal transfer outcomes on 128 held-out semantic sources. Because outcomes reuse models and queries, they are not independent trials. All observed transfers succeed, giving a pair/source-cluster bootstrap interval of $[1.000,1.000]$, which does not guarantee accuracy on unseen sources. Wrong-target donor packets yield $0\%$ accuracy, while idealized one-hot semantic packets yield $93.75\%$, indicating task sensitivity and recipient dependence on packet realization. These results concern two training objectives within the selected query-conditioned architecture family.

\paragraph{Image identity partitioning and query reuse.}
The CIFAR-100 test split contains 5,000 admission and 5,000 held-out image records, balanced at 50 per class per partition; exact raw-pixel duplicates remain together even when their labels differ. This prevents identical-pixel overlap across partitions but not historical exposure to official test imagery during method development. Final evaluation uses 5,000 queries sampled with replacement from held-out records and shared across all 64 ordered model pairs, enabling paired comparisons while introducing dependence. A 1,000-replicate bootstrap resamples donors, recipients, and shared-query positions; with four models per architecture, its intervals remain descriptive. The primary endpoint is $g_{\mathrm{swap}}$, reported separately by cross-architecture direction, without importing the prospective study's non-inferiority margin or fitting adjustments to held-out outcomes.

\paragraph{Complementary CIFAR-100 evaluation questions.}
The amended prospective 20-model study tests an analytic-reference non-inferiority criterion, whereas the separate eight-model identity-disjoint study measures recipient-native substitution. Their model populations, image-access histories, endpoints, and resampling procedures differ, so their numerical results should not be pooled or treated as independent replications of one effect. They provide complementary evidence under their respective evaluation conditions.

\subsection{Additional Clean-Room Compatibility Results}
\label{app:structured-onboarding}

\paragraph{Structured nonterminal controls.}
A 16-valued joint-state task $(a',b')$ preserves the withheld-continuation requirement and yields a clean-minus-control target-probability advantage of $0.744$ (95\% CI $[0.708,0.780]$) across 2,450 directed pairs, with positive-edge density $1.0$. Removing the explicit $4+4$ coordinate partition also preserves nonterminal exchange: a fixed dense public decoder yields clean-donor target probability $0.506$ versus $0.037$ under corruption, a contrast of $0.469$ (95\% CI $[0.443,0.496]$), with every evaluated pair positive and public joint-state decoding accuracy $1.0$. These results show that the tested continuation effect does not require the original factor-specific coordinate partition.

\textbf{Additional protocol realizations preserve prospective compatibility.} A separately initialized 50-model later cohort is trained under the structured 8D protocol fixed before its training. Direct transplantation yields clean-versus-control contrasts of $0.751$ (95\% CI $[0.720,0.780]$) for future$\rightarrow$incumbent exchange and $0.748$ ($[0.720,0.776]$) for incumbent$\rightarrow$future exchange, with positive-edge density $1.0$ in both directions.

Table~\ref{tab:cleanroom-summary} summarizes later-cohort interoperability across the continuous-memory and dense structured-packet realizations.

\textbf{Prospective compatibility also survives the dense-decoder realization.} Without the explicit $4+4$ coordinate partition, but with a fixed public dense decoder, later-cohort transplantation achieves clean-versus-control contrasts of $0.461$ (95\% CI $[0.438,0.485]$) for future$\rightarrow$incumbent exchange and $0.459$ ($[0.433,0.485]$) for incumbent$\rightarrow$future exchange, with positive-edge density $1.0$ in both directions. This supports prospective compatibility without dedicated factor coordinates in the tested realization.

\begin{table}[t]
\centering
\caption{\textbf{Later-cohort interoperability across additional protocol realizations.}
Direct transplantation within and across model generations. Effects are corrupted-minus-clean MSE for continuous memory and clean-minus-control target-probability advantages for dense structured packets; parentheses report 95\% model-node bootstrap confidence intervals. Positive-edge density is $1.0$ throughout.}
\label{tab:cleanroom-summary}
\scriptsize
\setlength{\tabcolsep}{3.1pt}
\begin{tabular}{llr@{\hspace{4pt}}l}
\toprule
Study & Direction & Pairs & Effect (95\% CI) \\
\midrule
\multirow{3}{*}{Continuous memory}
& Incumbent$\rightarrow$Incumbent & 870 & 0.291 (0.286, 0.296) \\
& Future$\rightarrow$Inc. & 900 & 0.291 (0.287, 0.296) \\
& Inc.$\rightarrow$Future & 900 & 0.291 (0.286, 0.295) \\
\midrule
\multirow{3}{*}{Dense structured}
& Incumbent$\rightarrow$Incumbent & 2,450 & 0.469 (0.443, 0.496) \\
& Future$\rightarrow$Inc. & 2,500 & 0.461 (0.438, 0.485) \\
& Inc.$\rightarrow$Future & 2,500 & 0.459 (0.433, 0.485) \\
\bottomrule
\end{tabular}
\end{table}

\textbf{Representational structure and prospective compatibility.} The structured-state controls demonstrate nonterminal exchange without dedicated factor coordinates, while the later-cohort experiments demonstrate direct exchange with models trained after the public interface was fixed. Together, they support prospective compatibility across the tested protocol realizations without requiring a particular factor-specific coordinate layout.
\subsection{Continuous-Memory Experimental Details}
\label{app:continuous-details}

\textbf{Continuous-memory design.} We train 30 incumbent and 30 later models from independent random initializations, with the public protocol and primary analysis fixed before training. Each packet is a four-token memory $Z\in\mathbb{R}^{4\times16}$; a fixed dense public head evaluates its semantics through permutation-invariant matching of tokens to the underlying facts. For donor $i$, recipient $j$, and evaluation example $x$, squared prediction error is
\[
e^{\kappa}_{ijx}=(\hat y^{\kappa}_{ijx}-y_x)^2,
\qquad
\kappa\in\{\mathrm{clean},\mathrm{ctrl}\}.
\]

\textbf{Primary transfer endpoint.} For each ordered donor--recipient pair, we define
\[
\Delta^{(\mathrm{MSE})}_{ij}
=
\mathbb{E}_x\!\left[
e^{\mathrm{ctrl}}_{ijx}
-
e^{\mathrm{clean}}_{ijx}
\right].
\]
The matched control changes only the continuous value associated with the queried key, so $\Delta^{(\mathrm{MSE})}_{ij}>0$ indicates lower prediction error with the clean transferred memory than with the control. The positive-edge threshold is therefore zero, and the prespecified population criterion requires positive-edge density of at least $0.8$. Evaluation uses 2{,}048 fixed held-out examples. Analyses introduced after specification of the primary endpoint are labeled post hoc and are used only to probe robustness or scope.

\textbf{Continuous memories transfer prospectively in both directions.} All 30 incumbent and 30 later-trained models satisfy native-task and same-model interface qualification under the fixed public protocol. Across 870 directed incumbent pairs, the clean-versus-control MSE advantage is $0.291$ (95\% CI $[0.286,0.296]$), with positive-edge density $1.0$. Cross-generation exchange yields advantages of $0.291$ ($[0.287,0.296]$) for future$\rightarrow$incumbent and $0.291$ ($[0.286,0.295]$) for incumbent$\rightarrow$future, with positive-edge density $1.0$ in both directions.

\paragraph{Additional post-hoc analyses.}
\label{app:continuous-controls}
\textbf{Direct-semantic control.} We replace each learned 16D memory token with its canonical 12D semantic fact token, retaining the consumer architecture and training procedure apart from the input projection. All 30 models pass same-model interface qualification, and mean clean MSE is $2.92\times10^{-4}$, versus $0.008485$ for incumbent cross-model transfer using naturally emitted learned packets. This provides an empirical semantic-performance reference, not an information-matched competing method.

\textbf{Decoder-space sensitivity.} Let $W\in\mathbb{R}^{12\times16}$ be the bias-free public decoder, $W^{\dagger}$ its Moore--Penrose pseudoinverse, and $P=W^{\dagger}W\in\mathbb R^{16\times16}$ the orthogonal row-space projector. For token $z$, define
\[
z_{\mathrm{row}}=Pz,\qquad
z_{\mathrm{null}}=(I_{16}-P)z.
\]
The decoder-null component accounts for $1.49\%$ of mean token energy. Row-space projection increases clean MSE to $0.0153$, while matched-norm decoder-null randomization increases it to $0.0266$, despite decoded-semantic changes below $9\times10^{-6}$. Clean-minus-control contrasts remain positive at $0.291$ and $0.285$, respectively, with positive-edge density $1.0$. These synthetic interventions demonstrate recipient sensitivity to decoder-invisible packet components without establishing that the modified packets satisfy all protocol-validity requirements.

\textbf{Naturally emitted nullspace variation can be tolerated across donors.} We replace each token's decoder-null component with one naturally emitted by a different donor and matched to the same semantic fact, excluding self-swaps. Across 2{,}048 held-out examples and 870 off-diagonal donor--recipient pairs, decoded semantics change by at most $5.7\times10^{-6}$, while MSE changes from $0.008485$ to $0.008427$. The swap-minus-original difference is $-5.8\times10^{-5}$ (95\% CI $[-6.4\times10^{-4},\,5.1\times10^{-4}]$), with positive-edge density $1.0$. Unlike synthetic decoder-null perturbations, these matched natural swaps show no detectable deterioration, supporting tolerance of the tested variation rather than invariance to arbitrary nullspace changes.

\textbf{Evaluation-set robustness.} Across three independently sampled held-out test sets, interoperability effects remain stable for incumbent and later-model exchange in both directions. This supports robustness to the tested evaluation draws without establishing generalization to other data distributions.
\subsection{Implementation Variation (EQ6, EQ10)}
\label{app:implementation-variation}

\textbf{Protocol boundaries should be defined by computational role rather than literal layer address.} In the four-block depth variant, retaining the layer-specific address \texttt{B2.R1[QUERY]} yields mean same-model recovery of $0.685$, with a minimum of $0.190$. Using the role-defined boundary \texttt{FINAL.R1[QUERY]} instead supports reliable same-model recovery across the evaluated depth and width variants, as well as cross-implementation transfer. The negative control shows that a fixed layer address need not preserve the computational function of an exchange boundary.

\textbf{Architecture qualification and full-population evaluation.} The one-layer GRU fails to meet task and same-model interface competence criteria consistently, even with extended training: 36/50 and 44/50 models qualify in two runs. Its cross-family transfer results therefore cannot isolate interoperability from insufficient model competence. A two-layer GRU with hidden size 32 qualifies in 48/50 cases, but we retain all 50 models in the transfer analysis. Across this unfiltered population, Transformer$\to$GRU advantage is $0.832$ (95\% CI $[0.802,0.862]$) and GRU$\to$Transformer advantage is $0.790$ (95\% CI $[0.768,0.813]$), with positive-edge density $1.0$ in both directions and no pair-specific fitting. These results support cross-architecture transfer in the evaluated population without implying that all 50 GRUs satisfy qualification criteria.

\subsection{Exploratory Frozen-Consumer Compatibility Controls}\label{app:dcc-matched}
\paragraph{Question and evaluation.}
We examine frozen-consumer substitution under a public-semantic packet objective and two alternative CIFAR-100 training constructions. These development-only diagnostics are not matched comparisons with the original compatible-representation method of \citet{gygli2021reusable}. Evaluations use the same manifest-defined 2,048 incumbent-only consumer-fitting images, 5,000 development images, and 20,000 continuation queries. The DCC interface has 128 dimensions and a 49,472-parameter learned consumer, versus 64 dimensions and 33,088 parameters for NIP. Because interface dimensionality and consumer capacity differ, numerical contrasts do not isolate the training objective.

\paragraph{Frozen-consumer compatibility under shared-classifier training.}
We test whether training with a shared learned classifier yields frozen-consumer interchangeability. An independently trained ResNet-18 incumbent (20,000 updates) and ConvNeXt entrant (30,000 updates), both using batch size 128, optimize continuation MSE and shared-classifier cross-entropy with coefficient 1. The entrant reuses the incumbent's learned classifier head, which is excluded from entrant optimizer updates and verified unchanged after training. Their respective 2.56 and 3.84 million image exposures match the corresponding NIP training budgets. The incumbent-fitted consumer achieves native recovery $0.3238$, exceeding the provisional $0.30$ competence threshold. Substituting entrant features reduces recovery to $0.1932$, with entrant-relative excess MSE $0.1308$ (incumbent MSE $0.6773$; entrant MSE $0.8081$). This single-pair adaptation does not reproduce the full method of \citet{gygli2021reusable}, and qualifying the incumbent consumer does not establish matched native accuracy or supervision across methods.

\paragraph{NIP recipe contrast.}
Across three separately trained, development-only ResNet-to-ConvNeXt pairs, entrant-relative excess MSE is $0.0287$, $0.0069$, and $0.0427$ (mean $0.0261$) for incumbent-fitted NIP consumers, versus $0.1232$, $0.0785$, and $0.0930$ for the corresponding public-target compatibility control (paired mean advantage for NIP $0.0721$). The control both removes the canonical-consumer penalty and reallocates its weight to the task loss; this is a comparison of training recipes, not a causal ablation of that penalty. The separate shared-classifier control involves one pair rather than three and uses a different consumer interface, so the unpaired comparison supports no between-pair significance claim. These exploratory results do not alter the prespecified $0.03$ analytic-reference margin in the 200-edge prospective CIFAR-100 study.

\paragraph{Interpretation of exploratory compatibility diagnostics.}
The frozen-consumer experiments examine how selected training objectives and shared-head constructions affect substitution of representations across implementations. The earlier classification-only comparison showed substantial degradation with an incumbent-fitted ridge semantic readout, whereas the joint-task comparison uses a separately fitted, native-competent consumer. Differences in continuation supervision, consumer fitting, and semantic readout prevent a controlled ranking of compatibility methods.

Neither experiment reproduces the full method of \citet{gygli2021reusable}, and the unpaired NIP-versus-shared-classifier results do not establish general method-level superiority. Together, these exploratory diagnostics highlight the importance of recipient competence, interface dimensionality, information access, and matched objectives when evaluating interoperability.

\section{Formal Details}
\label{app:formal}

\textbf{Formal foundations of specification-first neural interoperability.} This appendix formalizes the semantically indexed Neural Interoperability Protocols used in our experiments. It defines the public packet space, semantic validity sets, producer--consumer obligations, and clean-room independence, distinguishing universal conformance requirements from evaluations on naturally emitted packets. It then gives a sufficient condition for pair-independent compatibility, characterizes private reparameterizations that preserve public-interface behavior, and distinguishes decoder consistency from full packet validity. These formal results are conditional on the stated assumptions and do not establish universal conformance of the evaluated models.

\subsection{Task factorization and protocol specialization}

\textbf{Task factorization.} Let $\mathcal U$, $\mathcal C$, and $\mathcal Y$ denote the producer-input, continuation, and output spaces. We assume an intermediate semantic state $S$ satisfying
\[
S=s(U),
\qquad
Y=F(S,C),
\]
where $s:\mathcal U\to\mathcal S$ and $F:\mathcal S\times\mathcal C\to\mathcal Y$, with $F$ defined on the full product $\mathcal S\times\mathcal C$. This factorization provides a semantic reference for intermediate-state exchange in the evaluated tasks; it does not assert that arbitrary neural tasks admit a unique or identifiable decomposition. In the compositional settings, $C$ is unavailable at packet emission. The recipient must therefore combine the transferred state with the subsequently supplied continuation, making the packet an intermediate computational resource rather than an encoding of the final answer.

\textbf{Protocol specialization.} For the semantically indexed family studied here, define
\[
\Pi=\bigl(\mathcal Z,\mathcal E,\{\mathcal Z_s\}_{s\in\mathcal S},\beta,\tau\bigr).
\]
Here $\mathcal Z$ is the public packet space; $\mathcal A$ denotes the supported private-state or architecture classes, with public exporters $\mathcal E=\{E_a\}_{a\in\mathcal A}$; and $\mathcal Z_s\subseteq\mathcal Z$ contains the protocol-valid packets representing semantic state $s$. The boundary specification $\beta$ identifies the computational role of packet emission and consumption, while $\tau$ defines the public transplantation rule. These objects specify a donor--recipient-independent representation contract. Architecture-specific exporters define permitted interfaces without requiring access to other implementations' learned representations.

\textbf{Packet geometry and direct transplantation.} The public packet space $\mathcal Z$ is protocol-specific, with both vector-valued and sequence-valued realizations in our experiments. In each realization,
\[
\tau(z)=z,
\]
meaning that donor packets are supplied to recipients unchanged in public coordinates, without pair-specific alignment, adapters, or calibration. Compatibility therefore concerns conformance to a shared public packet contract rather than learning mappings between private representations.

\textbf{Public exporters map private states into the common packet space.} For each supported implementation class $a$,
\[
E_a:\mathcal H_a\to\mathcal Z,
\]
where $\mathcal H_a$ denotes its private state space, assumed to be a finite-dimensional real vector space for the coordinate transformations considered below. In the evaluated constructions, $E_a$ is public, deterministic, and fixed independently of participating models. It may depend on declared state shape or architecture class but is not learned from incumbent or future private representations. Thus the common packet space is specified without fitting an alignment to the models being connected. Determinism is a property of these constructions, not a requirement of the abstract NIP definition.

\textbf{Participant factorization.} Implementation $j$ has a private upstream map
\[
H_j:\mathcal U\to\mathcal H_{a(j)},
\]
a private importer
\[
I_j:\mathcal Z\to\mathcal R_j,
\]
and private downstream computation, distinct from the public semantic decoder or readout,
\[
D_j:\mathcal R_j\times\mathcal C\to\mathcal Y.
\]
The resulting computation is
\[
u
\xrightarrow{H_j}
h_j
\xrightarrow{E_{a(j)}}
z_j
\xrightarrow{I_j}
r_j
\xrightarrow{D_j(\cdot,c)}
\hat y_j.
\]
The public exporter maps private state into $\mathcal Z$, and the importer maps the packet into recipient-private state. Although $I_j$ may be learned within implementation $j$, it is not fitted to a particular donor. Compatibility depends on satisfying the public packet contract rather than learning donor-specific mappings between private representations.

\textbf{Exact canonicalization is sufficient but not required.} If
\[
\mathcal Z_s=\{\phi(s)\},
\]
all conforming producers emit the same public packet for semantic state $s$. This point-canonicalized regime supports interoperability when recipients satisfy their corresponding obligations, but a protocol need not require unique packet representations. More generally, $\mathcal Z_s$ may contain multiple valid packets for the same semantic state, provided conforming recipients correctly consume every packet in that set.

\paragraph{Scope of the validity sets.}
For a deterministic consumer to satisfy strong conformance, any states $s,t$ sharing a protocol-valid packet must require identical outputs for every continuation:
\[
\mathcal Z_s\cap\mathcal Z_t\neq\varnothing
\quad\Longrightarrow\quad
F(s,c)=F(t,c)
\quad\text{for all }c\in\mathcal C.
\]
Otherwise, the consumer would be required to produce different outputs from the same packet and continuation.

\textbf{Declared packet validity.} For each semantic state $s$, the set $\mathcal Z_s$ defines its protocol-valid public packets. These validity sets are specified independently of the packets emitted by trained models and may admit multiple representations of the same state. However, agreement under a public semantic decoder does not necessarily establish packet validity: a packet belongs to $\mathcal Z_s$ only if it satisfies the declared requirements, which may extend beyond decoder output. The following definitions specify universal conformance obligations; their sampled experimental counterparts are distinguished below.

\paragraph{Definition 1 (producer conformance).}
Implementation $j$ satisfies producer conformance if
\[
E_{a(j)}(H_j(u))\in\mathcal Z_{s(u)}
\]
for every valid producer input $u$.

\paragraph{Definition 2 (strong consumer conformance).}
Implementation $j$ satisfies strong consumer conformance if
\[
D_j(I_j(z),c)=F(s,c)
\]
for every $s\in\mathcal S$, every $z\in\mathcal Z_s$, and every $c\in\mathcal C$.

\textbf{Two complementary interface obligations.} Producer conformance requires every valid input to yield a packet in the declared validity set for its semantic state. Strong consumer conformance requires correct continuation for every packet in that validity set and every permitted continuation. These obligations are distinct: sampled producer checks do not establish universal conformance, and successful transfer from tested donors does not establish consumer correctness over the full valid-packet domain. Strong consumer conformance is a sufficient formal condition, not a property established by the experiments.

\paragraph{Compatibility versus native-recipient performance.}
Definition~2 requires correctness relative to $F(s,c)$ on protocol-valid packets, not equality of expected loss between imperfect learned producers or preservation of a recipient's own-packet performance under substitution. The identity-disjoint image study separately measures this empirical property using the same recipient and sampled queries:
\[
g_{\mathrm{swap}}
=
R_{\mathrm{recipient\ clean}}-R_{\mathrm{transfer}},
\]
where $R$ denotes normalized squared-error recovery. This gap measures the observed substitution effect, not universal consumer conformance over the full validity sets. A negative gap indicates higher recovery with the transferred packet and need not imply a recipient defect.

\subsection{Clean-room compatibility under a fixed protocol}

Let the public protocol $\Pi$ be fixed before training the future population $\mathcal M_{\mathrm{future}}$. Future implementations may use the same public task resources and protocol supervision as other implementations trained against $\Pi$, but have no access to incumbent checkpoints, parameters, hidden states, activations, naturally emitted packets, learned importers, pair-specific losses or transformations, alignment data, incumbent-derived protocol targets, or other incumbent-private representation information.

Thus, clean-room independence excludes incumbent-private representation information, not shared public task resources.

Future population $\mathcal M_{\mathrm{future}}$ is clean-room compatible with incumbent population $\mathcal M_{\mathrm{inc}}$ if it satisfies the defined future$\rightarrow$incumbent transplantation criterion; bidirectional compatibility additionally requires incumbent$\rightarrow$future transfer.

The structured-state extension is exploratory. In contrast, the dSprites clean-room and primary continuous-memory studies fix their protocols and analyses before training the evaluation populations. These results concern compatibility with later models trained against a previously specified public protocol, not arbitrary models that were never trained to implement it.

\subsection{Natural packet distributions and empirical evaluation}

\paragraph{Empirical versus strong conformance.}
Producer objectives and public-decoder diagnostics assess semantic agreement on sampled inputs, not universal producer conformance. Cross-model transplantation tests recipient behavior on naturally emitted donor packets rather than every packet in $\mathcal Z_s$. Matched corruption controls support causal interpretation of the measured transfer effects but do not establish behavior on arbitrary decoder-consistent packets. This distinction matters when the decoder has a nontrivial nullspace: correct decoded semantics do not rule out recipient sensitivity to decoder-invisible components.

For a trained producer population $\mathcal M$, let $Q_s^{\mathcal M}$ denote its distribution of naturally emitted packets for semantic state $s$. This distribution depends on the population and training procedure; it does not define the protocol-valid set $\mathcal Z_s$. Successful transfer on $Q_s^{\mathcal M}$ therefore need not imply strong consumer conformance over $\mathcal Z_s$. The experiments instead assess causal use of naturally emitted packets from independently trained models under their specified transplantation endpoints.

\paragraph{Mapping formal obligations to empirical checks.}
Finite evaluations test selected consequences of the public producer and consumer obligations, not universal conformance. Producer-side diagnostics compare emitted packets with public semantic targets on held-out inputs; any additional packet-validity requirements need separate verification. Recipient-side transplantation measures $D_j(I_j(z_i),c)$ using naturally emitted packets $z_i$ from independent donors and sampled continuations $c$, with matched corruption controls where specified. Neither diagnostic covers every valid input, packet, or continuation. We therefore distinguish \emph{strong consumer conformance} (Definition~2) from \emph{observed interoperability} under the evaluated populations, qualification criteria, packet distributions, and endpoints. Decoder-nullspace interventions reveal recipient sensitivity but cannot establish counterexamples to strong conformance unless the modified packets are independently shown to satisfy the full protocol-validity requirements.

\subsection{Compatibility from independent conformance}

\textbf{Compatibility from shared conformance obligations.} Proposition~1 establishes sufficient conditions for pair-independent compatibility: a conforming producer emits protocol-valid packets, and a strongly conforming recipient correctly consumes every packet allowed by the same specification. Under these conditions, direct exchange requires no donor--recipient-specific transformation. The result follows from the formal conformance obligations; it does not establish that any particular training method satisfies them.

\paragraph{Proposition 1 (pair-independent compatibility).}
If donor $i$ satisfies producer conformance and recipient $j$ satisfies strong consumer conformance to the same protocol, then
\[
D_j\!\left(
I_j(E_{a(i)}(H_i(u))),
c
\right)
=
F(s(u),c)
\]
for every valid producer input $u$ and every $c\in\mathcal C$, without a donor--recipient-specific transformation.

\paragraph{Proof.}
Producer conformance implies
\[
E_{a(i)}(H_i(u))
\in
\mathcal Z_{s(u)},
\]
so strong consumer conformance applies directly.
\hfill$\square$

Proposition~1 gives sufficient conditions for pair-independent compatibility: producer conformance and correct recipient continuation over the declared validity sets. Under these conditions, direct transplantation requires no pair-specific alignment. The result does not establish that training achieves these conditions, that finite evaluation can certify them, or that strong conformance is computationally tractable to verify. It is therefore a conditional specification-level guarantee, not a representation-learning guarantee.

\paragraph{Reading the held-out evidence against the proposition.}
In structured retrieval, implementations trained with different semantic objectives achieve exact task success when exchanging naturally emitted packets, consistent with Proposition~1 but not sufficient to verify its universal premises. The identity-disjoint CIFAR-100 study instead measures graded recovery on sampled image identities: its positive ResNet-to-ConvNeXt and negative ConvNeXt-to-ResNet substitution gaps show that passing finite qualification criteria does not ensure performance invariance under donor replacement. Neither result contradicts Proposition~1, which assumes strong consumer conformance over every protocol-valid packet. High canonical-packet recovery and successful sampled transplantation cannot establish that premise. Formal compatibility, observed direct transfer, and lossless substitution at a specified endpoint therefore remain distinct claims.

\subsection{Private representational freedom}

Public compatibility does not require unique private coordinates. Let the post-import state space $\mathcal R_j$ be a finite-dimensional real vector space. For any invertible linear map $Q_j\in\mathrm{GL}(\mathcal R_j)$, define
\[
I'_j(z)=Q_jI_j(z),
\qquad
D'_j(r,c)=D_j(Q_j^{-1}r,c).
\]
Then
\[
D'_j(I'_j(z),c)
=
D_j(I_j(z),c).
\]

For producer-side transformations, define
\[
\mathcal G_{E_a}
=
\{T\in\mathrm{GL}(\mathcal H_a):E_a\circ T=E_a\},
\]
where $\mathrm{GL}(\mathcal H_a)$ denotes the invertible linear maps on the private state space. Every $T\in\mathcal G_{E_a}$ preserves the exported public packet, even if it changes private coordinates.

\paragraph{Proposition 2 (private reparameterization invariance).}
Conformance is invariant to private invertible reparameterizations that preserve the public packet and are compensated within private downstream computation. Hence, public interoperability does not imply a unique global private coordinate system.

\paragraph{Proof.}
For the recipient transformation, $Q_j^{-1}Q_j$ cancels inside $D_j$, so $D'_j(I'_j(z),c)=D_j(I_j(z),c)$ for every packet $z$ and continuation $c$. On the producer side, $E_a(T(h))=E_a(h)$ for every $T\in\mathcal G_{E_a}$ and private state $h$, so emitted packets are unchanged. These are compensated private coordinate changes, not a claim of invariance under arbitrary uncoordinated transformations.\hfill$\square$

\textbf{Interoperability without identical private representations.} Proposition~2 shows that public-interface behavior is preserved under compensated invertible changes of recipient-private coordinates or producer-side transformations that leave exported packets unchanged. Thus a shared public interface does not require identical private representations. This is a standard change-of-coordinates property, not a new symmetry theorem or a characterization of all neural-network symmetries. The phrase \emph{partial gauge fixing at the interface} is descriptive: the public contract constrains exchanged representations without fixing private coordinates uniquely. The partial-contract and decoder-nullspace experiments probe related aspects of interface behavior but do not establish invariance under arbitrary private transformations.

\subsection{Decoder consistency versus protocol validity}

The dense-decoder structured-state experiment illustrates the distinction between semantic-decoder consistency and full packet validity. Identify each semantic state $s$ with a class index in $\{1,\ldots,16\}$, and let
\[
W_{\mathrm{pub}}\in\mathbb R^{16\times8}
\]
be the fixed public decoder. Define the decoder-consistent region
\[
\widetilde{\mathcal Z}_s
=
\left\{
z\in\mathbb R^8:
\operatorname*{arg\,max}_{\ell\in\{1,\ldots,16\}}(W_{\mathrm{pub}}z)_\ell=s
\right\},
\]
where $\operatorname*{arg\,max}$ returns a single index, breaking exact ties in favor of the smallest class index.

Decoder consistency is a public constraint but need not fully characterize protocol validity. In the evaluated decoder-anchored realization, every protocol-valid packet for state $s$ must satisfy the public decoder check. This explicit requirement, which is not imposed by the abstract NIP definition, gives
\[
\mathcal Z_s
\subseteq
\widetilde{\mathcal Z}_s.
\]
The inclusion may be strict: a decoder-consistent packet need not satisfy every requirement for protocol validity.

The experiment demonstrates successful direct transplantation of naturally emitted packets between independently trained implementations despite continuous within-class variation. It does not establish recipient correctness over the full decoder-consistent region $\widetilde{\mathcal Z}_s$.

\textbf{Shared interface semantics with representational flexibility.} The decoder-consistency analysis shows that direct exchange can succeed despite variation beyond the public semantic constraint. In the evaluated implementations, naturally emitted packets support successful recipient continuation, but decoder-consistent alternatives need not satisfy the full protocol. These results establish neither convergence to a unique latent representation nor recipient invariance over the entire decoder-consistent region.

\textbf{Formal synthesis.} The framework separates three questions that should not be conflated: which obligations a public interface declares, which behaviors follow when implementations satisfy those obligations, and which of those behaviors have been observed experimentally. Proposition~1 establishes direct compatibility under producer and strong-consumer conformance; Proposition~2 identifies compensated private-coordinate transformations that preserve interface behavior. The distinction between decoder-consistent packets and declared validity sets further explains why successful natural-packet exchange does not certify strong consumer conformance. Together, these statements formalize the scope of a specification-first approach without requiring identical private representations or claiming guarantees beyond the stated premises.

\section{AI Use Disclosure}\label{app:ai-use}

Generative AI tools assisted with drafting, editing, literature organization, code review, and
presentation.
The author reviewed all generated material and takes responsibility for the scientific claims,
experimental design, implementation, analysis, citations, and final manuscript.
AI tools were not treated as autonomous authors or as substitutes for experimental validation.

\end{document}